\documentclass{article}
\usepackage[preprint]{neurips_2020}
\usepackage[utf8]{inputenc}
\usepackage[T1]{fontenc}
\usepackage[english]{babel}
\usepackage[colorlinks]{hyperref}
\usepackage{url}
\usepackage{amsfonts}
\usepackage{nicefrac}
\usepackage{microtype}
\usepackage{amsmath}
\usepackage{amssymb}
\usepackage{mathtools}
\usepackage{subcaption}
\usepackage{booktabs}
\usepackage{ragged2e}
\usepackage{tikz}
\usepackage{stackengine}
\usepackage{etoolbox}
\usepackage{xspace}
\usepackage{xpatch}
\usepackage{cuted}
\usepackage{enumerate}
\usepackage{xstring}
\usepackage{natbib}
\usepackage{setspace}
\usepackage{tabularx}
\usepackage{makecell}
\usepackage{graphicx}
\usepackage{changepage}
\usepackage{enumitem}
\usepackage{cuted}
\usepackage{titlesec}
\usepackage{cancel}
\usepackage{eqparbox}
\usepackage{wrapfig}
\usepackage[hang,flushmargin]{footmisc}
\usepackage[capitalise,noabbrev,nameinlink]{cleveref}
\usepackage[useregional=numeric]{datetime2}
\usepackage[linesnumbered,ruled,noend]{algorithm2e}
\usepackage{tocbasic}
\usepackage{eqparbox}
\usepackage{amssymb}
\usepackage{pifont}
\usepackage{titlecaps}

\usetikzlibrary{positioning,arrows.meta,fit,calc,backgrounds,shapes.geometric}
\pgfdeclarelayer{background}
\pgfdeclarelayer{behind}
\pgfdeclarelayer{above}
\pgfsetlayers{background,behind,main,above}
\tikzset{%
node distance=2em, auto,
every node/.style={line width=0.7pt},
det/.style={diamond, draw=black, fill=white, minimum size=2.5em, inner sep=0.1ex},
lat/.style={circle, draw=black, fill=white, minimum size=2.5em, inner sep=0.1ex},
obs/.style={circle, draw=black, fill=black!15, minimum size=2.5em, inner sep=0.1ex},
fac/.style={rectangle, draw=black, fill=black, minimum size=.6em, inner sep=0em},
param/.style={},
dummy/.style={circle, draw=none, minimum size=2.5em},
plate/.style={rounded corners, draw=black, fill=white, inner sep=.8em, align=right},
box/.style={rounded corners, draw=black, inner sep=.4em, align=center},
backg/.style={rounded corners, fill=black!6},
generates/.style={->, -{Stealth[length=.6em, inset=0pt]}, line width=0.7pt},
undirected/.style={line width=0.7pt},
}

\setlist[itemize]{leftmargin=1.3em,itemsep=0em,topsep=.1em}
\makeatletter
\renewcommand{\paragraph}[1]{\textbf{#1}\quad\@ifnextchar\par\@gobble\relax}
\makeatother
\DeclareTOCStyleEntry[beforeskip=.2em plus 1pt]{tocline}{section}
\xapptocmd\normalsize{%
\abovedisplayskip=.7em plus .2em minus .2em
\belowdisplayskip=.3em plus .1em minus .1em
\abovedisplayshortskip=.7em plus .2em minus .2em
\belowdisplayshortskip=.3em plus .1em minus .1em
}{}{}

\newcommand{\itempar}[1]{\item\textbf{#1}\quad}

\makeatletter
\DeclareDocumentCommand\todo{g}{%
\def\@message{\IfNoValueTF{#1}{TODO}{TODO: #1}}
\textbf{\textcolor[HTML]{FF8811}{\@message}}
\@latex@warning{\@message}{}{}}
\makeatother

\Addlcwords{and as but for if nor or so yet a an the as at by for in of off on per to up via}
\setcitestyle{authoryear,round,citesep={;},aysep={,},yysep={;}}
\renewcommand{\cite}[1]{\citep{#1}}
\creflabelformat{equation}{#2\textup{#1}#3}
\crefname{algocf}{Algorithm}{Algorithms}
\Crefname{algocf}{Algorithm}{Algorithms}

\definecolor{mydarkblue}{rgb}{0,0.08,0.45}
\hypersetup{%
colorlinks=true,
linkcolor=mydarkblue,
citecolor=mydarkblue,
filecolor=mydarkblue,
urlcolor=mydarkblue}

\graphicspath{{figures/}}

\renewcommand\d{\mathop{}\!\textnormal{\slshape d}}

\newcommand\dotprop{\overset{\smash{\raisebox{-.5ex}{\ensuremath{\cdot}}}}{\ensuremath{\propto}}}

\newcommand{\describe}[3][0pt]{\hspace*{.12em}\underbracket[0.5pt][1pt]{#2\hspace*{#1}}_\text{\clap{#3}}}

\newcommand{\textlabel}[1]{\text{#1:}\quad}
\newcommand{\Sum}{\textstyle\sum}
\newcommand{\Prod}{\textstyle\prod}

\makeatletter
\newcommand{\removeParBefore}{\ifvmode\vspace*{-\baselineskip}\setlength{\parskip}{0ex}\fi}
\newcommand{\removeParAfter}{\@ifnextchar\par\@gobble\relax}
\newcommand{\eq}{\begingroup\removeParBefore\endlinechar=32 \eqinner}
\newcommand{\eqinner}[2][aligned]{\endlinechar=32%
\begin{gather}\begin{#1}#2\end{#1}\end{gather}\endgroup\removeParAfter}
\makeatother

\DeclareDocumentCommand{\p}{ D<>{p} D<>{} D(){} }{\ensuremath{
\def\content{#3}\patchcmd{\content}{|}{\;#2\vert\;}{}{}
#1 \ifdefempty{\content}{}{#2(\content #2)}}}

\DeclareDocumentCommand{\P}{ D<>{P} D<>{\big} r() }{
\def\content{#3}\patchcmd{\content}{|}{\;#2\vert\;}{}{}
\ensuremath{\operatorname{#1}#2(\content #2)}}

\DeclareDocumentCommand{\E}{ D<>{E} E{_}{{}} D<>{\big} r[] }{
\def\content{#4}\patchcmd{\content}{|}{\;#3\vert\;}{}{}
\ensuremath{\operatorname{#1}_{#2}\!#3[\content #3]}}

\DeclareDocumentCommand{\D}{ D<>{D} D<>{\big} r[] }{
\def\content{#3}\patchcmd{\content}{||}{\;#2\|\;}{}{}
\ensuremath{\operatorname{#1}\!#2[\content #2]}}

\NewDocumentCommand{\Nor}{ r() }{\P<Normal>](#1)}
\NewDocumentCommand{\Cat}{ r() }{\P<Cat>](#1)}
\NewDocumentCommand{\Bin}{ r() }{\P<Bin>](#1)}
\NewDocumentCommand{\Bet}{ r() }{\P<Beta>](#1)}
\NewDocumentCommand{\Ber}{ r() }{\P<Bernoulli>(#1)}
\NewDocumentCommand{\Dir}{ r() }{\P<Dir>(#1)}

\DeclareDocumentCommand{\KL}{ D<>{\big} r[] }{\D<KL><#1>[#2]}
\DeclareDocumentCommand{\EKL}{ D<>{\big} r[] }{\D<E\,KL><#1>[#2]}
\DeclareDocumentCommand{\H}{ D<>{\big} E{_}{{}} r[] }{\E<H>_{#2}<#1>[#3]}
\DeclareDocumentCommand{\I}{ D<>{\big} E{_}{{}} r[] }{\E<I>_{#2}<#1>[#3]}

\DeclareDocumentCommand{\pp}{ D<>{} D(){} }{\ensuremath{\p<p_\phi><#1>(#2)}}
\DeclareDocumentCommand{\lnp}{ D<>{} D(){} }{\ensuremath{\p<\ln p><#1>(#2)}}
\DeclareDocumentCommand{\lnpp}{ D<>{} D(){} }{\ensuremath{\p<\ln p_\phi><#1>(#2)}}
\DeclareDocumentCommand{\q}{ D<>{} D(){} }{\ensuremath{\p<\tau><#1>(#2)}}
\DeclareDocumentCommand{\lnq}{ D<>{} D(){} }{\ensuremath{\p<\ln \tau><#1>(#2)}}
\DeclareDocumentCommand{\qp}{ D<>{} D(){} }{\ensuremath{\p<\tau_\phi><#1>(#2)}}
\DeclareDocumentCommand{\lnqp}{ D<>{} D(){} }{\ensuremath{\p<\ln \tau_\phi><#1>(#2)}}

\title{Action and Perception as Divergence Minimization}
\author{%
Danijar Hafner \\ Google Brain \And
Pedro A. Ortega \\ DeepMind \And
Jimmy Ba \\ University of Toronto \AND
Thomas Parr \\ University College London \And
Karl Friston \\ University College London \And
Nicolas Heess \\ DeepMind}

\begin{document}

\maketitle

\vfill

\begin{abstract}
\begin{hyphenrules}{nohyphenation}
To learn directed behaviors in complex environments, intelligent agents need to optimize objective functions. Various objectives are known for designing artificial agents, including task rewards and intrinsic motivation. However, it is unclear how the known objectives relate to each other, which objectives remain yet to be discovered, and which objectives better describe the behavior of humans. We introduce the Action Perception Divergence (APD), an approach for categorizing the space of possible objective functions for embodied agents. We show a spectrum that reaches from narrow to general objectives. While the narrow objectives correspond to domain-specific rewards as typical in reinforcement learning, the general objectives maximize information with the environment through latent variable models of input sequences. Intuitively, these agents use perception to align their beliefs with the world and use actions to align the world with their beliefs. They infer representations that are informative of past inputs, explore future inputs that are informative of their representations, and select actions or skills that maximally influence future inputs. This explains a wide range of unsupervised objectives from a single principle, including representation learning, information gain, empowerment, and skill discovery. Our findings suggest leveraging powerful world models for unsupervised exploration as a path toward highly adaptive agents that seek out large niches in their environments, rendering task rewards optional.

% We introduce a unified objective for action and perception of intelligent agents. Extending representation learning and control, we minimize the joint divergence between the combined system of agent and environment and a target distribution. Intuitively, such agents use perception to align their beliefs with the world, and use actions to align the world with their beliefs. Minimizing the joint divergence to an expressive target maximizes the mutual information between the agent's representations and inputs, thus inferring representations that are informative of past inputs and exploring future inputs that are informative of the representations. This lets us explain intrinsic objectives, such as representation learning, information gain, empowerment, and skill discovery from minimal assumptions. Moreover, interpreting the target distribution as a latent variable model suggests powerful world models as a path toward highly adaptive agents that seek large niches in their environments, rendering task rewards optional. The framework provides a common language for comparing a wide range of objectives, advances the understanding of latent variables for decision making, and offers a recipe for designing novel objectives. We recommend deriving future agent objectives the joint divergence to facilitate comparison, to point out the agent's target distribution, and to identify the intrinsic objective terms needed to reach that distribution.

\end{hyphenrules}
\end{abstract}

\vfill
\begin{figure}[h!]
\centering
\vspace*{2ex}
\hspace*{-2ex}
\includegraphics[width=.9\linewidth]{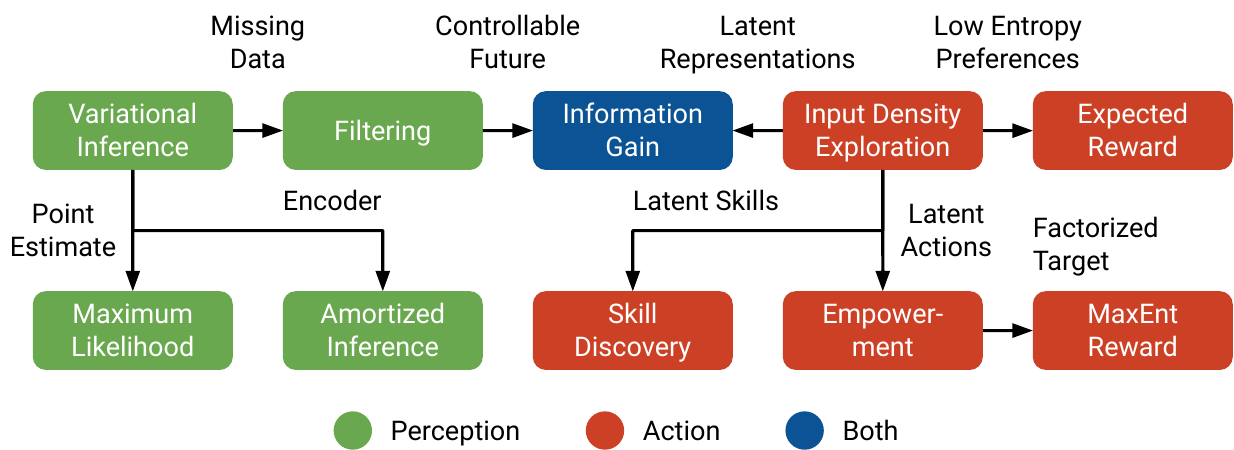}
\vspace*{-.5ex}
\caption{Overview of methods connected by the introduced framework of action and perception as divergence minimization. Each latent variable leads to a mutual information term between said variable and the data. The mutual information with past inputs explains representation learning. The mutual information with future inputs explains information gain, empowerment, and skill discovery. By leveraging multiple latent variables for the decision making process, agents can naturally combine multiple of the objectives. This figure shows the methods that drive from the well-established KL divergence and analogous method trees can be derived by choosing different divergence measures.}
\label{fig:overview}
\end{figure}
\vfill
\vfill
\clearpage

\section{Introduction}
\label{sec:intro}

To achieve goals in complex environments, intelligent agents need to perceive their environments and choose effective actions. These two processes, perception and action, are often studied in isolation. Despite the many objectives that have been proposed in the fields of representation learning and reinforcement learning, it remains unclear how the objectives relate to each other and which fundamentally new objectives remain yet to be discovered. Based on the KL divergence \citep{kullback1951kl}, we propose a unified framework for action and perception that connects a wide range of objectives to facilitate our understanding of them while providing a recipe for designing novel agent objectives. Our findings are conceptual in nature and this paper includes no empirical study. Instead, we offer a unified picture of a wide range of methods that have been shown to be successful in practice in prior work. The contributions of this paper are described as follows.

\paragraph{Unified objective function for perception and action}

We propose joint KL minimization as a principled framework for designing and comparing agent objectives. KL minimization was proposed separately for perception as variational inference \citep{jordan1999vi,alemi2018therml} and for actions as KL control \citep{todorov2008duality,kappen2009klcontrol}. Based on this insight, we formulate action and perception as jointly minimizing the KL from the world to a unified target distribution. The target serves both as the model to infer representations and as reward for actions. This extends variational inference to controllable inputs, while extending KL control to latent representations. We show a novel decomposition of joint KL divergence that explains several representation learning and exploration objectives. Divergence minimization additionally connects deep reinforcement learning to the free energy principle \citep{friston2010fep,friston2019physics}, while simplifying and overcoming limitations of its active inference implementations \citep{friston2017activeinference} that we discuss in \cref{sec:actinf}.

\paragraph{Understanding latent variables for decision making}

Divergence minimization with an expressive target maximizes the mutual information between inputs and latents. Agents thus infer representations that are informative of past inputs and explore future inputs that are informative of the representations. For the past, this yields reconstruction \citep{hinton2006deepbelief,kingma2013vae} or contrastive learning \citep{gutmann2010nce,oord2018cpc}. For the future, it yields information gain exploration \citep{lindley1956expectedinfo}. Stochastic skills and actions are realized over time, so their past terms are constant. For the future, they lead to empowerment \citep{klyubin2005empowerment} and skill discovery \citep{gregor2016vic}. RL as inference \citep{rawlik2010controlasinference} does not maximize mutual information because its target is factorized. To optimize a consistent objective across past and future, latent representations should be accompanied by information gain exploration.

\paragraph{Expressive world models for large ecological niches}

The more flexible an agent's target or model, the better the agent can adapt to its environment. Minimizing the divergence between the world and the model, the agent converges to a natural equilibrium or niche where it can accurately predict its inputs and that it can inhabit despite external perturbations \citep{schrodinger1944life,wiener1948cybernetics,haken1981synergetics,friston2013life,berseth2019smirl}. While surprise minimization can lead to trivial solutions, divergence minimization encourages the niche to match the agent's model class, thus visiting all inputs proportionally to how well they can be understood. This suggests designing expressive world models of sensory inputs \citep{ebert2017visualmpc,hafner2018planet,gregor2019beliefstate} as a path toward building highly adaptive agents, while rendering task rewards optional. 

\section{Framework}
\label{sec:framework}

This section introduces the Action Perception Divergence (APD). To unify action and perception, we formulate the two processes as joint KL minimization with a shared target distribution. The target distribution expresses the agent's preferences over system configurations and is also the probabilistic model under which the agent infers its representations. Using an expressive model as the target maximizes the mutual information between the latent variables and the sequence of sensory inputs, thus inferring latent representations that are informative of past inputs and exploring future inputs that are informative of the representations. We assume knowledge of basic concepts from probability and information theory that are reviewed in \cref{sec:background}.

\subsection{Joint KL Minimization}
\label{sec:global}

Consider a stochastic system described by a joint probability distribution over random variables. For example, the random variables for supervised learning are the inputs and labels and for an agent they are the sequence of sensory inputs, internal representations, and actions. More generally, we combine all input variables into $x$ and the remaining variables that we term latents into $z$. We will see that different latents correspond to different representation learning and exploration objectives.

The random variables are distributed according to their generative process or actual distribution $\pp$. Parts of the actual distribution can be unknown, such as the data distribution, and parts can be influenced by varying the parameter vector $\phi$, such as the distribution of stochastic representations or actions. As a counterpart to the actual distribution, we define the desired target distribution $\q$ over the same support. It describes our preferences over system configurations and can be unnormalized,

\eq{\text{Actual distribution:} \quad x,z\sim\pp(x,z) \quad\quad\quad\quad \text{Target distribution:} \quad \q(x,z).}

We formulate the problem of joint action and perception as moving the actual distribution of all random variables as close as possible to the target distribution, as measured by the KL divergence \citep{kullback1951kl,li2017alice,alemi2018therml}. To perform joint perception and action, the agent minimizes the Action Perception Divergence (APD) with respect to its parameter vector $\phi$,

\eq{\operatorname{APD}(\phi) \doteq \KL[\pp(x,z) || \q(x,z)]. \label{eq:jointkl}}

All expectations and KLs throughout the paper are integrals under the actual distribution, so they can be estimated from samples of the system and depend on $\phi$. \Cref{eq:jointkl} is the reverse KL or information projection used in variational inference \citep{csiszar2003projection}.

\paragraph{Examples}

For variational inference, $\pp$ is the joint of a fixed data distribution $\p(x)$ and the approximate posterior belief $\pp(z|x)$ and $\q(x,z)=\q(z)\q(x|z)$ is a latent variable model. Note that we use $\pp$ to denote not the model under which we infer beliefs but the generative process of inputs and their representations. For reinforcement learning, $\pp(x,z)$ is the trajectory distribution of inputs and actions under the current policy $\phi$ and $\q(x,z)=\exp(r(x))/Z$ is a Boltzmann distribution on the utility of the trajectory. The parameters $\phi$ include everything the optimizer can change directly, such as sufficient statistics of representations, model parameters, and policy parameters.

\paragraph{Target parameters}

There are two ways to denote deterministic values within our framework, also known as MAP estimates in the probabilistic modeling literature \citep{bishop2006book}. We can either have a fixed target distribution and use a latent variable that follows a point mass distribution \citep{dirac1958quantummechanics}, or we explicitly parameterize the target using a deterministic parameter as $\qp$. In either case, $\q$ refers to the fixed model class. The two approaches are equivalent because in both cases the target receives a deterministic value that has no entropy regularizer. For more details, see \cref{sec:vi}.

\paragraph{Assumptions}

Divergence minimization uses only two inductive biases, namely that the agent optimizes an objective and that it uses random variables to represent uncertainty. Choosing the well-established KL as the divergence measure is an additional assumption. It corresponds to maximizing the expected log probability under the target while encouraging high entropy for all variables in the system to avoid overconfidence, as detailed in \cref{sec:klinterp}.

\paragraph{Generality}

Alternative divergence measures would lead to different optimization dynamics, different solutions if the target cannot be reached, and potentially novel objectives for representation learning and exploration. Nonetheless, the KL can describe any converged system, trivially by choosing its actual distribution as the target, and thus offers a simple and complete mathematical perspective for comparing a wide range of specific objectives that correspond to different latent variables and target distributions.

\subsection{Information Bounds}
\label{sec:bounds}

We show that for expressive targets that capture dependencies between the variables in the system, minimizing the joint KL increases both the preferences and the mutual information between inputs $x$ and latents $z$. This property allows divergence minimization to explain a wide range of existing representation learning and exploration objectives. We use the term representation learning for inferring deterministic or stochastic variables from inputs, which includes local representations of individual inputs and global representations such as model parameters.

\paragraph{Latent preferences}

The joint KL can be decomposed in multiple ways, for example into a marginal KL plus a conditional KL or by grouping marginal with conditional terms. To reveal the mutual information maximization, we decompose the joint KL into a preference seeking term and an information seeking term. The decomposition can be done either with the information term expressed over inputs and the preferences expressed over latents or the other way around,

\eq{\describe{\KL[\pp(x,z) || \q(x,z)]}{joint divergence}
= \describe{\EKL[\pp(z|x) || \q(z)]}{realizing latent preferences}
- \describe{\E[\lnq(x|z)-\lnpp(x)]}{information bound}.
\label{eq:info_input}}

All expectations throughout the paper are over all variables, under the actual distribution, and thus depend on the parameters $\phi$. The first term on the right side of \cref{eq:info_input} is a KL regularizer that keeps the belief $\pp(z|x)$ over latent variables close to the marginal latent preferences $\q(z)$. The second term is a variational bound on the mutual information $\smash{\I[x;z]}$ \citep{barber2003variationalinfo}. The bound is expressed in input space. Maximizing the conditional $\ln\q(x|z)$ seeks latent variables that accurately predict inputs while minimizing the marginal $\ln\pp(x)$ seeks diverse inputs.

\paragraph{Variational free energy}

When the agent cannot influence its inputs, such as when learning from a fixed dataset, the input entropy $\E[-\ln\pp(x)]$ is not parameterized and can be dropped from \cref{eq:info_input}. This yields the expected free energy or ELBO objective used by variational inference to infer approximate posterior beliefs in latent variable models \citep{hinton1993vi,jordan1999vi}. It regularizes the belief $\pp(z|x)$ to stay close to the prior $\q(z)$ while reconstructing inputs via $\q(x|z)$. However, in reinforcement or active learning, input distribution changes and thus the input entropy should be kept unless the inputs are deterministic or homoskedastic.

\paragraph{Input preferences}

Analogously, we decompose the joint KL the other way around. The first term on the right side of \cref{eq:info_latent} is a KL regularizer that keeps the conditional input distribution $\pp(x|z)$ close to the marginal input preferences $\q(x)$. This term is analogous to the objective in KL control \citep{todorov2008duality,kappen2009klcontrol}, except that the inputs now depend upon latent variables via the policy. The second term is again a variational bound on the mutual information $\smash{\I[x;z]}$, this time expressed in latent space. Intuitively, the bound compares the belief $\q(z|x)$ after observing the inputs and the belief $\pp(z)$ before observing any inputs to measure the gained information,

\eq{\describe{\KL[\pp(x,z) || \q(x,z)]}{joint divergence}
= \describe{\EKL[\pp(x|z) || \q(x)]}{realizing input preferences}
- \describe{\E[\lnq(z|x)-\lnpp(z)]}{information bound}.
\label{eq:info_latent}}

The information bounds are tighter the better the target conditional approximates the actual conditional, meaning that the agent becomes better at maximizing mutual information as it learns more about the relation between the two variables. This requires an expressive target that captures correlations between inputs and latents, such as a latent variable model or deep neural network. Maximizing the mutual information accounts for both learning latent representations that are informative of inputs as well as exploring inputs that are informative of the latent representations.

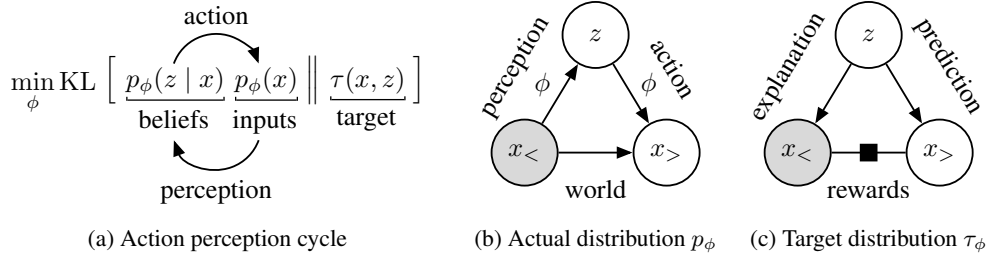
\begin{figure}[t]%
\vspace*{-3ex}%
\hspace*{1em}%
\begin{subfigure}[t]{.43\linewidth}
\centering
\begin{tikzpicture}[node distance=0em,eq/.style={inner sep=.1em},tx/.style={inner sep=.4em}]
\node[eq] (m) {$\displaystyle\min_\phi \operatorname{KL}$};
\node[eq,right=of m,yshift=.5em] (l) {$\Big[$};
\node[eq,right=of l,yshift=-.5em] (b) {$\describe{\pp(z|x)}{}$};
\node[eq,right=of b] (i) {$\describe{\pp(x)}{}$};
\node[eq,right=of i,yshift=.5em] (s) {$\Big\|$};
\node[eq,right=of s,yshift=-.5em] (t) {$\describe{\q(x,z)}{}$};
\node[eq,right=of t,yshift=.5em] (r) {$\Big]$};
\node[tx,below=of b,yshift=.9em] (bl) {beliefs};
\node[tx,below=of i,yshift=.9em] (il) {inputs};
\node[tx,below=of t,yshift=.9em] (tl) {target};
\draw[->, -{Stealth[length=.6em, inset=0pt]}, line width=0.7pt] (b.north)%
  arc[radius=.6, start angle=160, end angle=20] node[midway] {action};
\draw[<-, {Stealth[length=.6em, inset=0pt]}-, line width=0.7pt] (bl.south)%
  arc[radius=.6, start angle=200, end angle=340] node[midway,below] {perception};
\end{tikzpicture}%
\caption{Action perception cycle}
\label{fig:cycle}
\end{subfigure}\hfill%
\begin{subfigure}[t]{.23\linewidth}
\centering
\begin{tikzpicture}[node distance=2.8em, baseline=-5ex]
\clip (-0.6,-0.6) rectangle + (3.1,2.8);
\node[obs] (x) {$x_<$};
\node[lat, right=of x] (y) {$x_>$};
\node (d) at ($(x)!0.5!(y)$) {};
\node[lat, above=of d] (z) {$z$};
\path (x) edge[generates] node[yshift=-1ex] {$\phi$} node[sloped, anchor=south, inner sep=1.5em] {perception} (z);
\path (z) edge[generates] node[yshift=-1ex] {$\phi$} node[sloped, anchor=south, inner sep=1.5em] {action} (y);
\path (x) edge[generates] node[anchor=north, inner sep=1em] {world} (y);
\end{tikzpicture}
\caption{Actual distribution \pp}
\label{fig:actual_time}
\end{subfigure}\hfill%
\begin{subfigure}[t]{.23\linewidth}
\centering
\begin{tikzpicture}[node distance=2.8em, baseline=-5ex]
\clip (-0.6,-0.6) rectangle + (3.1,2.8);
\node[obs] (x) {$x_<$};
\node[lat, right=of x] (y) {$x_>$};
\node (d) at ($(x)!0.5!(y)$) {};
\node[lat, above=of d] (z) {$z$};
\node[fac] (f) at ($(x)!0.5!(y)$) {};
\path (z) edge[generates] node[sloped, anchor=south, inner sep=1.5em] {explanation} (x);
\path (z) edge[generates] node[sloped, anchor=south, inner sep=1.5em] {prediction} (y);
\path (x) edge[undirected] node[anchor=north, inner sep=1em] {rewards} (y);
\end{tikzpicture}
\caption{Target distribution \qp}
\label{fig:target_time}
\end{subfigure}%
\hspace*{1em}%
\caption{Action and perception minimize the joint KL divergence to a unified target distribution that can be interpreted as a learning probabilistic model of the system. Given the target, perception aligns the agent's beliefs with past inputs while actions align future inputs with its beliefs. There are many ways to specify the target, for example as a latent variable model that explains past inputs and predicts future inputs and an optional reward factor that is shown as a filled square.}
\label{fig:framework}
\end{figure}

\subsection{Models as Preferences}
\label{sec:preferences}

The target distribution defines our preferences over system configurations. However, we can also interpret it as a probabilistic model, or as an energy-based model if  it is unnormalized \citep{lecun2006energytutorial}. This is because minimizing the joint KL with under a fixed data distribution exactly recovers variational inference that infers approximate posteriors under the model. Moreover, as the agent brings the actual distribution closer toward the target, the target also becomes a better predictor of the actual system. By interpreting inference as bringing the joint of data and belief distributions toward the model, divergence minimization emphasizes that a model class simply expresses preferences over latent representations and inputs.

\paragraph{Action perception cycle}

Interpreting the target as a model shows that divergence minimization is consistent with the idea of perception as inference suggested by Helmholtz \citep{helmholtz1866perception,gregory1980perception}. Expressing preferences as models is inspired by the free energy principle and active inference \citep{friston2010fep,friston2012value,friston2017activeinference}, which we compare to in \cref{sec:actinf}. Divergence minimization inherits an interpretation of action and perception from active inference that we visualize in \cref{fig:cycle}. While action and perception both minimize the same joint KL, they affect different variables. Perception is based on inputs and affects the beliefs over representations, while actions are based on the representations and affect inputs. Given a unified target, perception thus aligns the agent's beliefs with the world while actions align the world with its beliefs.

\paragraph{Input preferences}

Minimizing the joint divergence also minimizes the divergence between the agent's input distribution $\pp(x)$ and the marginal input distribution under its target or model $\q(x)$. The marginal input distribution of the model is thus the agent's preferred input distribution, that the agent aims to sample from in the environment. Because the target can be unnormalized, we can combine a latent variable model with a reward factor of the form $\exp(r(x))$ to create an expressive target that incorporates task rewards. The reward factor shapes the input distribution without affecting the latents in the model. Alternatively, we can choose the target just as a reward factor, without a generative model, to recover normal reinforcement learning without explicit representation learning and information seeking. We make use of reward factors in \cref{sec:control,sec:maxentrl}.

\paragraph{Niche seeking}

The input preferences $\q(x)$ marginalizes out all latent variables and parameters and thus describe how well an input sequence $x$ can possibly be predicted by the model class \citep{jeffreys1935bayesfactor,kass1995bayesfactor}. This distribution may not be reachable due to environmental constraints. The agent thus converges to a natural equilibrium or niche in the environment that it can inhabit despite external perturbations \citep{wiener1948cybernetics,ashby1961cybernetics} and where it sees inputs proportionally to how well it can learn to understand them, while avoiding inputs that are inherently unpredictable given its model class. In deterministic or homoskedastic environments, the input entropy is constant, so that matching $\q(x)$ simplifies to surprise minimization \citep{schrodinger1944life,friston2013life,berseth2019smirl}, where the agent seeks input trajectories $x$ that maximize $\q(x)$. The information gain terms contributed by the latents prevent collapse to a trivial solution.

\paragraph{World models}

Choosing the target as a flexible model helps the agent reduce the joint divergence and thus better adapt to the environment or to changes in the environment. Expressive target distributions, such as world models, assign high probability to many input trajectories and thus result in large niches. Moreover, expressive targets lead to maximizing the mutual information between inputs and latents further, resulting in more informative representations and more effective information gain exploration. This suggests expressive world models as a path toward adaptive agents that understand and explore and inhabit large niches while rendering task rewards optional.

\subsection{Past and Future}
\label{sec:time}

Representations are computed from past inputs and exploration targets future inputs. To distinguish the two processes, we thus need to consider how an agent optimizes the joint KL after observing past inputs $x_<$ and before observing future inputs $x_>$, as discussed in \cref{fig:actual_time}. For example, past inputs can be stored in an experience dataset and future inputs can be approximated by either planning with a learned world model, on-policy trajectories, or replay of past inputs \citep{sutton1991dyna}. To condition the joint KL on past inputs, we first split the information bound from \cref{eq:info_input} into two smaller bounds on the past mutual information $\I[x_<;z]$ and additional future mutual information $\I[x_>;z|x_<]$,

\eq{\describe{\E[\lnq(z|x)-\lnpp(z)]}{information bound}
=&\E[\lnq(z|x)-\lnpp(z|x_<)+\lnpp(z|x_<)-\lnpp(z)\mkern3mu] \\[-2.5ex]
\geq&\E[\mkern-2mu\describe{\lnq(z|x)-\lnpp(z|x_<)}{future information bound}
      \mkern1mu+\describe{\lnq(z|x_<)\mkern7.5mu-\lnpp(z)}{past information bound}].
\label{eq:interm_belief}}

\Cref{eq:interm_belief} splits the belief update from the prior $\pp(z)$ to the posterior $\q(z|x)$ into two updates via the intermediate belief $\pp(z|x_<)$ and then applies the variational bound from \citet{barber2003variationalinfo} to allow both updates to be approximate. Splitting the information bound lets us separate past and future terms in the joint KL, or even separate individual time steps. It also lets us separately choose to express terms in input or latent space. This decomposition is one of our main contributions and shows how the joint KL divergence accounts for both representation learning and exploration,

\eq{\KL[\pp(x,z) || \q(x,z)] \leq
&\describe{\EKL[\pp(z|x_<) || \q(z)]}{realizing past latent preferences}
-\describe{\E[\lnq(x_<|z)-\lnpp(x_<)]}{representation learning} \\[-.3ex]
&\hspace{-6.2em}+\describe{\EKL[\pp(x_>|x_<,z) || \q(x_>|x_<)]}{realizing future input preferences}
-\describe{\E[\lnq(z|x)-\lnpp(z|x_<)]}{exploration} \\
&\hspace{-6.2em}=\text{maximize ELBO, reward, infogain, and empowerment}.
\label{eq:combined}}

Conditioning on past inputs $x_<$ removes their expectation and renders $\pp(x_<)$ constant. While some latent variables in the set $z$ are never realized, such as latent state estimates or model parameters, other latent variables become observed over time, such as stochastic actions or skills. Because the agent selects the values of these variables, we have to condition the objective terms on them as causal interventions \citep{pearl1995docalculus,ortega2010bayesiancontrolrule}. In practice, this means replacing all occurrences of $z$ by the unobserved latents $z_>$ and conditioning those terms on the observed latents $\operatorname{do}(z_<)$. To keep notation simple, we omit this step in our notation.

To build an intuition about \cref{eq:combined}, we discuss the four terms on the right-hand side. The first two terms involve the past while the last two terms involve the future. The first two terms are the ELBO objective for inferring representations, with $\pp(x_<)$ being a constant. The third term is the control objective that steers the agent toward the preferred input distribution $\q(x_>|x_<)$ that can be defined via a task reward. The fourth term is the expected information gain and empowerment, which encourages inputs that are informative of the agent's latents and encourages latents that have maximum influence over future inputs.

The bound in \cref{eq:combined} shows that agents that optimize representation learning and control terms separately, as shown on the right-hand side, optimize a bound on the joint KL. This suggests the promise of directly minimizing the joint KL end-to-end as opposed to learning representations only from past data but not to support future actions. The decomposition shows that the joint KL accounts for both learning informative representations of past inputs and exploring informative future inputs as two sides of the same coin. From this, we derive several representation and exploration objectives by including different latent variables in the set $z$. These objectives are summarized in \cref{tab:objectives} and derived with detailed examples in \cref{sec:examples}.

\begin{table}[t!]
\centering
\begin{tabular}{lcccl}
\toprule
\textbf{Latent} & \textbf{Target} & \textbf{Past Term} & \textbf{Future Term} & \textbf{Agents} \\
\midrule
Actions    & Factorized & ---            & Action entropy   & A3C, SQL, SAC \\
Actions    & Expressive & ---            & Empowerment      & VIM, ACIE, EPC \\
Skills     & Expressive & ---            & Skill discovery  & VIC, SNN, DIAYN, VALOR \\
States     & Expressive & Repr. learning & Information gain & NDIGO, DVBF-LM \\
Parameters & Expressive & Model learning & Information gain & VIME, MAX, Plan2Explore \\
% States     & Expressive & State estimation & Information gain & NDIGO, DVBF-LM \\
% Parameters & Expressive & System identification & Information gain & VIME, MAX, Plan2Explore \\
\bottomrule
\end{tabular}
\vspace*{1ex}
\caption{Divergence minimization accounts for a wide range of agent objectives. Each latent variable used by the agent contributes a future objective term. Moreover, latent variables that are not observed over time, such as latent representations and model parameters, additionally each contribute a past objective term. Combining multiple latent variables combines their objective terms. Refer to \cref{sec:examples} for detailed derivations of these individual examples and references for the listed agents.}
\label{tab:objectives}
\end{table}

\paragraph{Representation learning}

Because past inputs are observed, the past information bound only affects the latents. Expressed as \cref{eq:info_input}, it leads to reconstruction \citep{hinton2006deepbelief}, and as \cref{eq:info_latent}, it leads to contrastive learning \citep{gutmann2010nce,oord2018cpc}. This accounts for local representations of individual inputs, as well as global representations, such as latent parameters. Moreover, representations can be inferred online or amortized using an encoder \citep{kingma2013vae}. Latents with point estimates are equivalent to target parameters and thus are optimized jointly to tighten the variational bounds. Because past actions and skills are realized, their mutual information with realized past inputs is constant and thus contributes no past objective terms.

\paragraph{Exploration}

Under a flexible target, latents in $z$ result in information-maximizing exploration. For latent representations, this is known as expected information gain and encourages informative future inputs that convey the most bits about the latent variable, such as world model parameters, policy parameters, or state estimates \citep{lindley1956expectedinfo,sun2011plansurprise}. For stochastic actions, a fully factorized target leads to maximum entropy RL. An expressive target yields empowerment, maximizing the agent's influence over the world \citep{klyubin2005empowerment}. For skills, it yields skill discovery or diversity that learns distinct modes of behavior that together cover many different trajectories \citep{gregor2016vic,florensa2017snn,eysenbach2018diayn,achiam2018valor}.

\section{Unifying Review}
\label{sec:examples}

This section leverages the presented framework to explain a wide range of objectives in a unifying review, as outlined in \cref{fig:overview}. For this, we include different variables in the actual distribution, choose different target distributions, and then rewrite the joint KL to recover familiar objectives. We start with perception, the case with latent representations but uncontrollable inputs and then turn to action, the case without latent representations but with controllable inputs. We then turn to combined action and perception. The derivations follow the general recipe described in \cref{sec:framework}. The same steps can be followed for new latent structures and target distributions to yield novel agent objectives.

\subsection{Variational Inference}
\label{sec:vi}

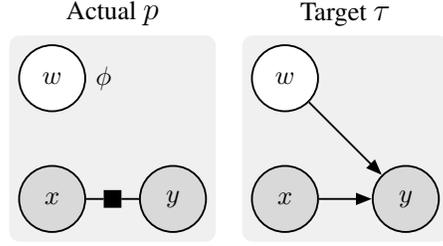
\begin{wrapfigure}{r}{.45\textwidth}%
\captionsetup{justification=centering}%
\vspace*{-6ex}%
\centering%
\begin{tikzpicture}
\node[obs] (x) {$x$};
\node[obs, right=of x] (y) {$y$};
\node[lat, param, above=of x] (w) {$w$};
\node[right=of w, xshift=-2em] (phi) {$\phi$};
\node[fac] (f) at ($(x)!0.5!(y)$) {};
\path (x) edge[undirected] node {} (f);
\path (y) edge[undirected] node {} (f);
\begin{scope}[on background layer] \node[backg,fit=(x)(y)(w)] {}; \end{scope}
\node[above] at (current bounding box.north) {Actual \large\p};
\end{tikzpicture}\hspace{1em}%
\begin{tikzpicture}
\node[obs] (x) {$x$};
\node[obs, right=of x] (y) {$y$};
\node[lat, above=of x] (w) {$w$};
\path (x) edge[generates] node {} (y);
\path (w) edge[generates] node {} (y);
\begin{scope}[on background layer] \node[backg,fit=(x)(y)(w)] {}; \end{scope}
\node[above] at (current bounding box.north) {Target \large\q};
\end{tikzpicture}%
\caption{Variational Inference}
\label{fig:vi}
\vspace*{-2ex}%
\end{wrapfigure}

Following Helmholtz, we describe perception as inference under a model \citep{helmholtz1866perception,gregory1980perception,dayan1995helmholtz}. Inference computes a posterior over representations by conditioning the model on inputs. Because this has no closed form in general, variational inference optimizes a parameterized belief to approximate the posterior \citep{peterson1987vi,hinton1993vi,jordan1999vi}.

\Cref{fig:vi} shows variational inference for the example of supervised learning using a BNN \citep{denker1987bnn,mackay1992bnn,blundell2015bbb}. The inputs are images $x\doteq\{x_i\}$ and their classes $y\doteq\{y_i\}$ and we infer the latent parameters $w$ as a global representation of the data set \citep{alemi2018therml}. The parameters depend on the inputs only through the optimization process that produces $\phi$. The target consists of a parameter prior and a conditional likelihood that uses the parameters to predict classes from images,

\eq{
&\textlabel{Actual} & \pp(x,y,w) & \doteq
  \describe{\pp(w)}{belief}
  \Prod_i \describe{\p(x_i,y_i)}{data}, \\
&\textlabel{Target} & \q(x,y,w) & \dotprop
  \describe{\q(w)}{prior}
  \Prod_i \describe{\q(y_i|x_i,w)}{likelihood}.}
  
Applying the framework, we minimize the KL between the actual and target joints. Because the data distribution is fixed here, the input marginal $\p(x,y)$ is a constant. In this case, the KL famously results in the free energy or ELBO objective \citep{hinton1993vi,jordan1999vi} that trades off remaining close to the prior and enabling accurate predictions. The objective can be interpreted as the description length of the data set under entropy coding \citep{huffman1952coding,mackay2003information} because it measures the nats needed for storing both parameter belief and prediction residuals,

\eq{\KL[\pp || \q] =
 \describe{\KL[\pp(w) || \q(w)]}{complexity}
-\describe{\E[\lnq(y|x,w)]}{accuracy}
+\describe{\E[\lnp(x,y)]}{constant}.}

Variational methods for BNNs \citep{peterson1987vi,hinton1993vi,blundell2015bbb} differ in their choices of prior and belief distributions and inference algorithm. This includes hierarchical priors \citep{louizos2016bnnmatrix,ghosh2017bnnhorseshoe}, data priors \citep{louizos2016bnnmatrix,hafner2018ncp,sun2019fbnn}, flexible posteriors \citep{louizos2016bnnmatrix,sun2017bnnstructure,louizos2017bnnflow,zhang2018naturalgradasvi,chang2019bnnensemble}, low rank posteriors \citep{izmailov2018bnnpca,dusenberry2020bnnrank1}, and improved inference algorithms \citep{wen2018flipout,immer2020bnnlocal}. BNNs have been leveraged for RL for robustness \citep{okada2020bayesianplanet,tran2019bayesianlayers} and exploration \citep{houthooft2016vime,azizzadenesheli2018bayesiandqn}.

\paragraph{Target parameters}

While expressive beliefs over model parameters lead to a global search for their values, provide uncertainty estimates for predictions, and enable directed exploration in the RL setting, they can be computationally expensive. When these properties are not needed, we can choose a point mass distribution $\pp(w) \rightarrow \delta_\phi(w) \doteq \{1\text{ if }w=\phi\text{ else } 0\}$ to simplify the expectations and avoid the entropy and mutual information terms that are zero for this variable \citep{dirac1958quantummechanics},

\begin{adjustwidth}{-1em}{-1em}
\eq{
 \describe{\KL[\pp(w) || \q(w)]}{complexity}
-\describe{\E[\lnq(y|x,w)]}{accuracy}
\rightarrow
 \describe{\lnq(\phi)}{complexity}
-\describe{\E[\lnq(y|x,\phi)]}{accuracy}
\doteq
 \describe{\E[-\lnqp(y|x)]}{parameterized target}.
\label{eq:map}}
\end{adjustwidth}

Point mass beliefs result in MAP or maximum likelihood estimates \citep{bishop2006book,murphy2012book} that are equivalent to parameterizing the target as $\qp$. Parameterizing the target is thus a notational choice for random variables with point mass beliefs. Technically, we also require the prior over target parameters to be integrable but this is true in practice where only finite parameter spaces exist.

\subsection{Amortized Inference}
\label{sec:amortized}

\begin{wrapfigure}{r}{.45\textwidth}%
\captionsetup{justification=centering}%
\vspace*{-6ex}%
\centering%
\begin{tikzpicture}
\node[obs] (x) {$x$};
\node[lat, above=of x] (z) {$z$};
\node[dummy, right=of x] (y) {};
\path (x) edge[generates] node {$\phi$} (z);
\begin{scope}[on background layer] \node[backg,fit=(x)(y)(z)] {}; \end{scope}
\node[above] at (current bounding box.north) {Actual \large\p};
\end{tikzpicture}\hspace{1em}%
\begin{tikzpicture}
\node[obs] (x) {$x$};
\node[lat, above=of x] (z) {$z$};
\node[dummy, right=of x] (y) {};
\path (z) edge[generates] node {$\phi$} (x);
\begin{scope}[on background layer] \node[backg,fit=(x)(y)(z)] {}; \end{scope}
\node[above] at (current bounding box.north) {Target \large\q};
\end{tikzpicture}%
\caption{Amortized Inference}
\label{fig:amortized}
\vspace*{-2ex}%
\end{wrapfigure}
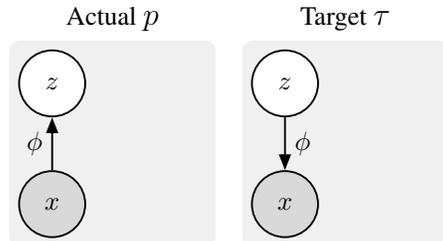

Local representations represent individual inputs. They can summarize inputs more compactly, enable interpolation between inputs, and facilitate generalization to unseen inputs. In this case, we can use amortized inference \citep{kingma2013vae,rezende2014vae,ha2016hypernets} to learn an encoder that maps each input to its corresponding belief. The encoder is shared among inputs to reuse computation. It can also compute beliefs for new inputs without further optimization, although optimization can refine the belief \citep{kim2018semiamortized}.

\Cref{fig:amortized} shows amortized inference on the example of a VAE \citep{kingma2013vae,rezende2014vae}. The inputs are images $x \doteq \{x_i\}$ and we infer their latent codes $z=\{z_i\}$. The actual distribution consists of the unknown and fixed data distribution and the parameterized encoder $\pp(z_i|x_i)$. The target is a probabilistic model defined as the prior over codes and the decoder that computes the conditional likelihood of each image given its code. We parameterize the target here, but one could also introduce an additional latent variable to infer a distribution over decoder parameters as in \cref{sec:vi},

\eq{
&\textlabel{Actual} & \pp(x,z) & \doteq
\Prod_i
  \describe{\p(x_i)}{data}
  \describe{\pp(z_i|x_i)}{encoder}, \\
&\textlabel{Target} & \qp(x,z) & \doteq
\Prod_i
  \describe{\qp(x_i|z_i)}{decoder}
  \describe{\q(z_i)}{prior}.}
  
Because the data distribution is still fixed, minimizing the joint KL again results in the variational free energy or ELBO objective that trades of prediction accuracy and belief simplicity. However, by including the constant input marginal, we highlight that the prediction term is a variational bound on the mutual information that encourages the representations to be informative of their inputs,

\eq{\KL[\pp || \qp]
=\describe{\EKL[\pp(z|x) || \q(z)]}{complexity}
-\describe{\E[\lnqp(x|z)-\lnp(x)]}{information bound}.}

In input space, the information bound leads to reconstruction as in DBNs \citep{hinton2006deepbelief}, VAEs \citep{kingma2013vae,rezende2014vae}, and latent dynamics \citep{krishnan2015deepkalman,karl2016dvbf}. In latent space, it leads to contrastive learning as in NCE \citep{gutmann2010nce}, CPC \citep{oord2018cpc,guo2018actioncpc}, CEB \citep{fischer2020ceb}, and SimCLR \cite{chen2020simclr}. To maximize their mutual information, $x$ and $z$ should be strongly correlated under the target distribution, which explains the empirical benefits of ramping up the decoder variance throughout learning \citep{bowman2015klannealing,eslami2018gqn} or scaling the temperature of the contrastive loss \citep{chen2020simclr}. The target defines the variational family and includes inductive biases \citep{tschannen2019miprior}. Both forms have enabled learning world models for planning \citep{ebert2018foresight,ha2018worldmodels,zhang2018solar,hafner2018planet,hafner2019dreamer} and accelerated RL \citep{lange2010rlrecon,jaderberg2016unreal,lee2019slac,yarats2019sacae,gregor2019beliefstate}.

\subsection{Future Inputs}
\label{sec:missing}

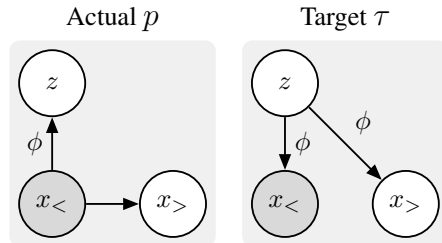
\begin{wrapfigure}{r}{.45\textwidth}%
\captionsetup{justification=centering}%
\vspace*{-6ex}%
\centering%
\begin{tikzpicture}
\node[obs] (x) {$x_<$};
\node[lat, right=of x] (y) {$x_>$};
\node[lat, above=of x] (z) {$z$};
\path (x) edge[generates] node {$\phi$} (z);
\path (x) edge[generates] node {} (y);
\begin{scope}[on background layer] \node[backg,fit=(x)(y)(z)] {}; \end{scope}
\node[above] at (current bounding box.north) {Actual \large\p};
\end{tikzpicture}\hspace{1em}%
\begin{tikzpicture}
\node[obs] (x) {$x_<$};
\node[lat, right=of x] (y) {$x_>$};
\node[lat, above=of x] (z) {$z$};
\path (z) edge[generates] node {$\phi$} (x);
\path (z) edge[generates] node {$\phi$} (y);
\begin{scope}[on background layer] \node[backg,fit=(x)(y)(z)] {}; \end{scope}
\node[above] at (current bounding box.north) {Target \large\q};
\end{tikzpicture}%
\caption{Future Inputs}
\label{fig:missing}
\vspace*{-2ex}%
\end{wrapfigure}

Before moving to actions, we discuss perception with unobserved future inputs that are outside of our control \citep{ghahramani1995missing}. This is typical in supervised learning where the test set is unavailable during training \citep{bishop2006book}, in online learning where training inputs become available over time \citep{amari1967onlinelearning}, and in filtering where only inputs up to the current time are available \citep{kalman1960filter}.

\Cref{fig:missing} shows missing inputs on the example of filtering with an HMM \citep{stratonovich1960hmm,kalman1960filter,karl2016dvbf}, although the same graphical model applies to supervied learning with a BNN or representation learning with a VAE given train and test data sets. The inputs $x \doteq \{x_<,x_>\}$ consist of past images $x_<$ and future images $x_>$ that follow an unknown and fixed data distribution. We represent the input sequence using a chain $z$ of corresponding compact latent states. However, the representations are computed only based on $x_<$ because $x_>$ is not yet available, as expressed in the factorization of the actual distribution,

\eq{
&\textlabel{Actual} & \pp(x,z) & \doteq
\describe{\p(x_>,x_<)}{data}\describe{\pp(z|x_<)}{belief}, \\
&\textlabel{Target} & \qp(x,z) & \doteq
\describe{\qp(x_<|z)}{likelihood}
\describe{\qp(x_>|z)}{prediction}
\describe{\q(z)}{prior}.}
  
\paragraph{Bayesian assumption}

Bayesian reasoning operates within the model class $\q$ and makes the assumption that the model class is correct. Under this assumption, the future inputs $x_> \sim \p(x_>|x_<,z)=\p(x_>|x_<)$ follow the target distribution $\qp(x_>|x_<,z)=\qp(x_>|z)$. This renders the divergence of future inputs given the other variables zero, so that $x_>$ does not need to be considered for optimization, recovering standard variational inference from \cref{sec:vi},

\eq{\KL[\pp || \qp]
= \describe{\KL[\pp(x_<,z) || \qp(x_<,z)]}{variational inference}
+ \describe{\EKL[\p(x_>|x_<) || \qp(x_>|z)]}{uncontrolled future}
\label{eq:missing_data}.}

Assuming that future inputs follow the model distribution is appropriate when the model accurately reflects our knowledge about future inputs. However, the assumption does not always hold, for example for data augmentation or distillation \citep{hinton2015distillation} that generate data from another distribution to improve the model. Importantly, assuming that future inputs already follow the target is not appropriate when they can be influenced, because there would be no need to intervene.

\subsection{Control}
\label{sec:control}

\begin{wrapfigure}{r}{.45\textwidth}%
\captionsetup{justification=centering}%
\vspace*{-6ex}%
\centering%
\begin{tikzpicture}
\node[obs] (x) {$x_<$};
\node[lat, right=of x] (y) {$x_>$};
\node[dummy, above=of x] (z) {};
\path (x) edge[generates] node {$\phi$} (y);
\begin{scope}[on background layer] \node[backg,fit=(x)(y)(z)] {}; \end{scope}
\node[above] at (current bounding box.north) {Actual \large\p};
\end{tikzpicture}\hspace{1em}%
\begin{tikzpicture}
\node[obs] (x) {$x_<$};
\node[lat, right=of x] (y) {$x_>$};
\node[dummy, above=of x] (z) {};
\node[fac] (f) at ($(x)!0.5!(y)$) {};
\path (x) edge[undirected] node {} (f);
\path (y) edge[undirected] node {} (f);
\begin{scope}[on background layer] \node[backg,fit=(x)(y)(z)] {}; \end{scope}
\node[above] at (current bounding box.north) {Target \large\q};
\end{tikzpicture}%
\caption{Control}
\label{fig:control}
\vspace*{-2ex}%
\end{wrapfigure}

We describe behavior as an optimal control problem where the agent chooses actions to move its distribution of sensory inputs toward a preference distribution over inputs that can be specified via rewards \citep{morgenstern1953seu,lee2019smm}. We first cover deterministic actions that lead to KL control \citep{kappen2009klcontrol,todorov2008duality} and input density exploration \citep{schmidhuber1991curiousmodel,bellemare2016cts,pathak2017icm}.

\Cref{fig:control} shows deterministic control with the input sequence $x \doteq \{x_t\}$ that the agent can partially influence by varying the parameters $\phi$ of the deterministic policy, control rule, or plan. In the graphical model, we group the input sequence into past inputs $x_<$ and future inputs $x_>$. There are no internal latent variables. The target describes the preferences over input sequences that can be unnormalized,

\eq{
&\textlabel{Actual} & \pp(x) & \doteq
\Prod_t \describe{\pp(x_t|x_{1:t-1})}{controlled dynamics}, \\
&\textlabel{Target} & \q(x) & \doteq
\Prod_t \describe{\q(x_t|x_{1:t-1})}{preferences}.}

Minimizing the KL between the actual and target joints maximizes log preferences and the input entropy. Maximizing the input entropy is a simple form of exploration known as input density exploration that encourages rare inputs and aims for a uniform distribution over inputs \cite{schmidhuber1991curiousmodel,oudeyer2007curiosity}. This differs from the action entropy of maximum entropy RL in \cref{sec:maxentrl} and information gain in \cref{sec:infogain} that takes inherent stochasticity into account,

\eq{\KL[\pp || \q] =
-\Sum_t\big(
  \describe{\E[\lnq(x_t|x_{1:t-1})]}{expected preferences}
 +\describe{\H[\pp(x_t|x_{1:t-1})]}{curiosity}\big).
\label{eq:control}}

\paragraph{Task reward}

Motivated by risk-sensitivity \citep{pratt1964risk,howard1972riskrl}, KL control \citep{kappen2009klcontrol} defines the preferences as exponential task rewards $\q(x_t|x_{1:t-1}) \dotprop \exp(r(x_t))$. KL-regularized control \citep{todorov2008duality} defines the preferences with an additional passive dynamics term $\q(x_t|x_{1:t-1}) \dotprop \exp(r(x_t))\p<\tau'>(x_t|x_{1:t-1})$. Expected reward \citep{sutton2018rlbook} corresponds to the preferences $\qp(x_t|x_{1:t-1}) \dotprop \exp(r(x_t))\pp(x_t|x_{1:t-1})$ that include the controlled dynamics. This cancels out the curiosity term in the joint KL, leading to a simpler objective that does not encourage rare inputs, which might limit exploration of the environment.

\paragraph{Input density exploration}

Under divergence minimization, maximizing the input entropy is not an exploration heuristic but an inherent part of the control objective. In practice, the input entropy is often estimated by learning a density model of individual inputs as in pseudo-counts \citep{bellemare2016cts}, latent variable models as in SkewFit \citep{pong2019skewfit}, unnormalized models as in RND \citep{burda2018rnd}, and non-parameteric models as in reachability \citep{savinov2018reachability}. More accurately, it can be estimated by a sequence model of inputs as in ICM \citep{pathak2017icm}. The expectation over inputs is estimated by sampling episodes from either the actual environment, a replay buffer, or a learned model of the environment \citep{sutton1991dyna}.

\subsection{Maximum Entropy RL}
\label{sec:maxentrl}

\begin{wrapfigure}{r}{.45\textwidth}%
\captionsetup{justification=centering}%
\vspace*{-6ex}%
% \vspace*{-3.5ex}%
\centering%
\begin{tikzpicture}
\node[obs] (x) {$ax_<$};
\node[lat, right=of x] (y) {$x_>$};
\node[lat, above=of y] (a) {$a_>$};
\path (x) edge[generates] node {$\phi$} (a);
\path (x) edge[generates] node {} (y);
\path (a) edge[generates] node {} (y);
\begin{scope}[on background layer] \node[backg,fit=(x)(y)(a)] {}; \end{scope}
\node[above] at (current bounding box.north) {Actual \large\p};
\end{tikzpicture}\hspace{1em}%
\begin{tikzpicture}
\node[obs] (x) {$ax_<$};
\node[lat, right=of x] (y) {$x_>$};
\node[lat, above=of y] (a) {$a_>$};
\node[fac] (f) at ($(x)!0.5!(y)$) {};
\path (x) edge[undirected] node {} (f);
\path (y) edge[undirected] node {} (f);
\begin{scope}[on background layer] \node[backg,fit=(x)(y)(a)] {}; \end{scope}
\node[above] at (current bounding box.north) {Target \large\q};
\end{tikzpicture}%
\vspace*{-.5ex}  % TODO
\caption{Maximum Entropy RL}
\label{fig:maxentrl}
\vspace*{-2ex}%
\end{wrapfigure}
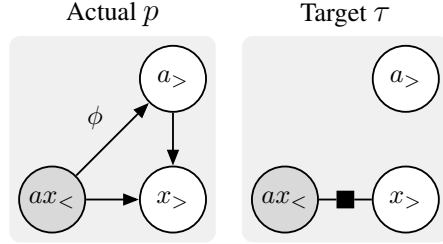

Instead of directly parameterizing the environment with a deterministic policy, we can treat actions as random variables through which the agent influences the environment. This results in stochastic policies that represent uncertainty over the best action for each situation. Stochastic policies allow exploration in action space \citep{williams1992reinforce,sutton2000policygradient,mnih2016a3c,haarnoja2018sac} and can be beneficial in environments with multiple learning agents \citep{bowling2000stochasticgames}.

\Cref{fig:maxentrl} shows stochastic control on the example of maximum entropy RL \citep{williams1991maxentreinforce,fox2015softq,schulman2017equivalence,haarnoja2017sql,haarnoja2018sac,levine2018maxent}. The input sequence is $x \doteq \{x_t\}$ and the action sequence is $a \doteq \{a_t\}$. In the graphical model, these are grouped into past actions and inputs $ax_<$, future actions $a_>$, and future inputs $x_>$. The actual distribution consists of the fixed environment dynamics and the stochastic policy. In maximum entropy RL, the target consists of a reward factor, an action prior that is often the same for all time steps, and the environment dynamics,

\eq{
&\textlabel{Actual} & \pp(x,a) & \doteq \Prod_t
  \describe{\p(x_t|x_{1:t-1},a_{1:t-1})}{environment}
  \describe{\pp(a_t|x_{1:t},a_{1:t-1})}{policy}, \\
&\textlabel{Target} & \q(x,a) & \dotprop \Prod_t
  \exp(\describe{r(x_t)}{reward})
  \describe{\p(x_t|x_{1:t-1},a_{1:t-1})}{environment}
  \describe{\q(a_t).}{action prior}}
  
Minimizing the joint KL results in a complexity regularizer in action space and the expected reward. Including the environment dynamics in the target cancels out the curiosity term as in the expected reward case in \cref{sec:control}, leaving maximum entropy RL to explore only in action space. Moreover, including the environment dynamics in the target gives up direct control over the agent's input preferences, as they depend not just on the reward but also the environment dynamics marginal. Because the target distribution is factorized and does not capture dependencies between $x$ and $a$, maximum entropy RL does not maximize their mutual information,

\eq{\KL[\pp || \q] = \Sum_t
  \describe{\EKL[\pp(a_t|x_{1:t},a_{1:t-1}) || \q(a_t)]}{complexity}
 -\describe{\E[r(x_t)].}{expected reward}
\label{eq:maxentrl}}

The action complexity KL can be simplified into an entropy regularizer by choosing a uniform action prior as in SQL \citep{haarnoja2017sql} and SAC \citep{haarnoja2018sac}. The action prior can also depend on the past inputs and incorporate knowledge from previous tasks as in Distral \citep{teh2017distral} and work by \citet{tirumala2019maxenttasks} and \citet{galashov2019defaultpolicy}. Divergence minimization motivates combining maximum entropy RL with input density exploration by removing the environment dynamics from the target distribution. The resulting agent aims to converge to the input distribution that is proportional to the exponentiated task reward.

\subsection{Empowerment}
\label{sec:empowerment}

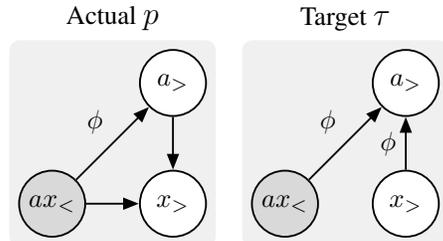
\begin{wrapfigure}{r}{.45\textwidth}%
\captionsetup{justification=centering}%
\vspace*{-6ex}%
\centering%
\begin{tikzpicture}
\node[obs] (x) {$ax_<$};
\node[lat, right=of x] (y) {$x_>$};
\node[lat, above=of y] (a) {$a_>$};
\path (x) edge[generates] node {$\phi$} (a);
\path (x) edge[generates] node {} (y);
\path (a) edge[generates] node {} (y);
\begin{scope}[on background layer] \node[backg,fit=(x)(y)(a)] {}; \end{scope}
\node[above] at (current bounding box.north) {Actual \large\p};
\end{tikzpicture}\hspace{1em}%
\begin{tikzpicture}
\node[obs] (x) {$ax_<$};
\node[lat, right=of x] (y) {$x_>$};
\node[lat, above=of y] (a) {$a_>$};
\path (x) edge[generates] node {$\phi$} (a);
\path (y) edge[generates] node {$\phi$} (a);
\begin{scope}[on background layer] \node[backg,fit=(x)(y)(a)] {}; \end{scope}
\node[above] at (current bounding box.north) {Target \large\q};
\end{tikzpicture}%
\caption{Empowerment}
\label{fig:empow}
\vspace*{-3ex}%
\end{wrapfigure}

Remaining in the stochastic control setting of \cref{sec:maxentrl}, we consider a different target distribution that predicts actions from inputs. This corresponds to an exploration objective that we term generalized empowerment, which maximizes the mutual information between the sequence of future inputs and future actions. It encourages the agent to influence its environment in as many ways as possible while avoiding actions that have no predictable effect.

\Cref{fig:empow} shows stochastic control with an expressive target that captures correlations between inputs and actions. The input sequence is $x \doteq \{x_t\}$ and the action sequence is $a \doteq \{a_t\}$. In the graphical model, these are grouped into past actions and inputs $ax_<$, future actions $a_>$, and future inputs $x_>$. The actual distribution consists of the environment and the stochastic policy. The target predicts actions from the inputs before and after them using a reverse predictor. We use uniform input preferences here, but the target can also include an additional reward factor as in \cref{sec:maxentrl},

\eq{
&\textlabel{Actual} & \pp(x,a) & \doteq \Prod_t
  \describe{\p(x_t|x_{1:t-1},a_{1:t-1})}{environment}
  \describe{\pp(a_t|x_{1:t},a_{1:t-1})}{policy}, \\
&\textlabel{Target} & \qp(x,a) & \dotprop \Prod_t
  \describe{\qp(a_t|x_{1:T},a_{1:t-1})}{reverse predictor}.}
  
Minimizing the joint KL reveals an information bound between future actions and inputs and a control term that maximizes input entropy and, if specified, task rewards. Empowerment \citep{klyubin2005empowerment} was originally introduced as potential empowerment to ``keep your options open'' and was later studied as realized empowerment to ``use your options'' \citep{salge2014realizedempow}. Realized empowerment maximizes the mutual information $\smash{\I[x_{t+k};a_{t:t+k}|x_{1:t},a_{1:t-1}]}$. Divergence minimization generalizes this to the mutual information $\smash{\I[x_{t:T};a_{t:T}|x_{1:t},a_{1:t-1}]}$ between the sequences of future actions and future inputs. The $k$-step variant is recovered by a target that conditions the reverse predictor on fewer inputs. Realized empowerment measures agent's influence on its environment and can be interpreted as maximizing information throughput with the action marginal $\pp(a_t|a_{t-1})$ as source, the environment as noisy channel, and the reverse predictor as decoder,

\eq[gathered]{\KL[\pp || \qp]
=\describe{\EKL[\p(x|a) || \q(x)]}{control}
 -\describe{\E[\lnqp(a|x)-\lnpp(a)]}{generalized empowerment}, \\
\describe{\E[\lnqp(a|x)-\lnpp(a)]}{generalized empowerment}
\geq \sum_t \E[
  \describe{\lnqp(a_t|x,a_{1:t-1})}{decoder}
 -\describe{\lnpp(a_t|a_{1:t-1})}{source}].
\label{eq:empowerment}}

Empowerment has been studied for continuous state spaces \citep{salge2013continuousempow}, for image inputs \citep{mohamed2015empowerment}, optimized using a variational bound \citep{karl2017empowerment}, combined with input density exploration \citep{de2018curiosityempow} and task rewards \citep{leibfried2019empowreward}, and used for task-agnostic exploration of locomotion behaviors \citep{zhao2020empow}. Divergence minimization suggests generalizing empowerment from the input $k$ steps ahead to the sequence of all future inputs. This can be seen as combining empowerment terms of different horizons. Moreover, we offer a principled motivation for combining empowerment with input density exploration. In comparison to maximum entropy RL in \cref{sec:maxentrl}, empowerment captures correlations between $x$ and $a$ in its target distribution and thus leads to information maximization. Moreover, it encourages the agent to converge to the input distribution that is proportional to the exponentiated reward.

\subsection{Skill Discovery}
\label{sec:skills}

\begin{wrapfigure}{r}{.45\textwidth}%
\captionsetup{justification=centering}%
\vspace*{-6ex}%
\centering%
\begin{tikzpicture}
\node[obs] (x) {$zax_<$};
\node[lat, above=of x] (z) {$z_>$};
\node[lat, right=of x] (y) {$x_>$};
\node[lat, above=of y] (a) {$a_>$};
\node[below=of z, yshift=2em] (phi) {$\phi$};
\path (x) edge[generates] node[xshift=.5em, yshift=.5em] {$\phi$} (a);
\path (x) edge[generates] node {} (y);
\path (z) edge[generates] node {} (a);
\path (a) edge[generates] node {} (y);
\begin{scope}[on background layer] \node[backg,fit=(x)(y)(a)] {}; \end{scope}
\node[above] at (current bounding box.north) {Actual \large\p};
\end{tikzpicture}\hspace{1em}%
\begin{tikzpicture}
\node[obs] (x) {$zax_<$};
\node[lat, above=of x] (z) {$z_>$};
\node[lat, right=of x] (y) {$x_>$};
\node[lat, above=of y] (a) {$a_>$};
\path (x) edge[generates] node {$\phi$} (z);
\path (y) edge[generates, right] node[yshift=.5em] {$\phi$} (z);
\begin{scope}[on background layer] \node[backg,fit=(x)(y)(a)] {}; \end{scope}
\node[above] at (current bounding box.north) {Target \large\q};
\end{tikzpicture}%
\caption{Skill Discovery}
\label{fig:skills}
\vspace*{-2ex}%
\end{wrapfigure}
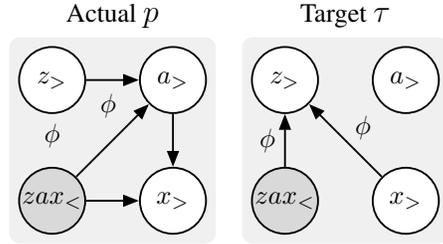

Many complex tasks can be broken down into sequences of simpler steps. To leverage this idea, we can condition a policy on temporally abstract options or skills \citep{sutton1999options}. Skill discovery aims to learn useful skills, either for a specific task or without rewards to solve downstream tasks later on. Where empowerment maximizes the mutual information between inputs and actions, skill discovery can be formulated as maximizing the mutual information between inputs and skills \citep{gregor2016vic}.

\Cref{fig:skills} shows skill discovery with the input sequence $x \doteq \{x_t\}$, action sequence $a \doteq \{a_t\}$, and the sequence of temporally abstract skills $z \doteq \{z_k\}$. The graphical model groups the sequences into past and future variables. The actual distribution consists of the fixed environment, an abstract policy that selects skills by sampling from a fixed distribution as shown here or as a function of past inputs, and the low-level policy that selects actions based on past inputs and the current skill. The target consists of an action prior and a reverse predictor for the skills and could further include a reward factor,

\eq{
&\textlabel{Actual} & \pp(x,a,z) & \doteq
  \Prod_{k=1}^{T/K}
    \describe{\pp(z_k)}{abstract policy}
  \Prod_{t=1}^T
    \describe{\pp(a_t | x_{1:t},a_{1:t-1},z_{\lfloor t/K \rfloor})}{policy}
    \describe{\p(x_t | x_{1:t-1}, a_{1:t-1})}{environment}, \\
&\textlabel{Target} & \qp(x,a,z) & \dotprop
  \Prod_{k=1}^{T/K}
    \describe{\qp(z_k | x)}{reverse predictor}
  \Prod_{t=1}^T
    \describe{\q(a_t)}{action prior}.
\raisetag{6ex}}

Minimizing the joint KL results in a control term as in \cref{sec:empowerment}, a complexity regularizer for actions as in \cref{sec:maxentrl}, and a variational bound on the mutual information between the sequences of inputs and skills. The information bound is a generalization of skill discovery \citep{gregor2016vic,florensa2017snn}. Conditioning the reverse predictor only on inputs that align with the duration of the skill recovers skill discovery. Maximizing the mutual information between skills and inputs encourages the agent to learn skills that together realize as many different input sequences as possible while avoiding overlap between the sequences realized by different skills,

\eq{\KL[\pp || \qp] =
\describe{\EKL[\p(x|a) || \tau(x)]}{control}
+\describe{\EKL[\pp(a|x,z) || \tau(a)]}{complexity}
-\describe{\E[\lnqp(z|x) - \lnpp(z)]}{skill discovery}.}

VIC \citep{gregor2016vic} introduced information-based skill discovery as an extension of empowerment, motivating a line of work including SNN \citep{florensa2017snn}, DIAYN \citep{eysenbach2018diayn}, work by \citet{hausman2018skills}, VALOR \citep{achiam2018valor}, and work by \citet{tirumala2019maxenttasks} and \citep{shankar2020temporalvi}. DADS \citep{sharma2019dads} estimates the mutual information in input space by combining a forward predictor of skills with a contrastive bound. Divergence minimization suggests a generalization of skill discovery where actions should not just consider the current skill but also seek out regions of the environment where many skills are applicable.

\subsection{Information Gain}
\label{sec:infogain}

\begin{wrapfigure}{r}{.45\textwidth}%
\captionsetup{justification=centering}%
\vspace*{-6ex}%
\centering%
\begin{tikzpicture}
\node[obs] (x) {$x_<$};
\node[lat, right=of x] (y) {$x_>$};
\node[lat, above=of x] (w) {$w$};
\path (x) edge[generates] node {$\phi$} (y);
\node[right=of w, xshift=-2em] (phi) {$\phi$};
\begin{scope}[on background layer] \node[backg,fit=(x)(y)(w)] {}; \end{scope}
\node[above] at (current bounding box.north) {Actual \large\p};
\end{tikzpicture}\hspace{1em}%
\begin{tikzpicture}
\node[obs] (x) {$x_<$};
\node[lat, right=of x] (y) {$x_>$};
\node[lat, above=of x] (w) {$w$};
\path (w) edge[generates,left] node {} (x);
\path (w) edge[generates] node {} (y);
\begin{scope}[on background layer] \node[backg,fit=(x)(y)(w)] {}; \end{scope}
\node[above] at (current bounding box.north) {Target \large\q};
\end{tikzpicture}%
\caption{Information Gain}
\label{fig:infogain}
\vspace*{-2ex}%
\end{wrapfigure}
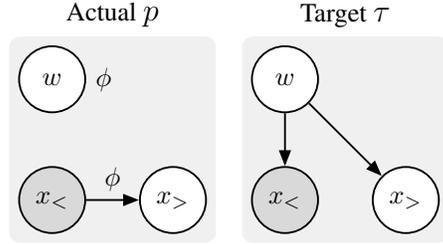

Agents need to explore initially unknown environments to achieve goals. Learning about the world is beneficial even when it does not serve maximizing the currently known reward signal, because the knowledge might become useful later on during this or later tasks. Reducing uncertainty requires representing uncertainty about aspects we want to explore, such as dynamics parameters, policy parameters, or state representations. To efficiently reduce uncertainty, the agent should select actions that maximize the expected information gain \citep{lindley1956expectedinfo}.

\Cref{fig:infogain} shows information gain exploration on the example of latent model parameters and deterministic actions. The inputs are a sequence $x \doteq \{x_t\}$ and the latent parameters are a global representation $w$. The graphical model separates inputs into past inputs $x_<$ and future inputs $x_>$. The actual distribution consists of the controlled dynamics and the parameter belief. Amortized latent state representations would include a link from $x_<$ to $z$. Latent policy parameters would include a link from $w$ to $x_>$. The target distribution is a latent variable model that explains past inputs and predicts future inputs, as in \cref{sec:missing}. The target could further include a reward factor,

\eq{
&\textlabel{Actual} & \pp(x,w) & \doteq
\describe{\pp(w)}{belief}
\Prod_t \describe{\pp(x_t|x_{1:t-1})}{controlled dynamics}, \\
&\textlabel{Target} & \q(x,w) & \doteq
\describe{\q(w)}{prior}
\Prod_t \describe{\q(x_t|x_{1:t-1},w)}{likelihood}.}

Minimizing the KL between the two joints reveals a control term as in previous sections and the information bound between inputs and the latent representation, as derived in \cref{sec:bounds}. In contrast to \cref{sec:missing}, we can now influence future inputs. This leads to learning representations that are informative of past inputs and exploring future inputs that are informative of the representations. The mutual information between the representation and future inputs is the expected information gain \citep{lindley1956expectedinfo,mackay1992infogain} that encourages inputs that are expected to convey the most bits about the representation to maximally reduce uncertainty in the belief,

\eq{\KL[\pp || \qp] \leq
 &\describe{\EKL[\pp(w|x_<) || \q(w)]}{simplicity}
- \describe{\E[\lnqp(x_<|w)-\lnpp(x_<)]}{representation learning} \\
+&\describe{\EKL[\pp(x_>|x_<,w) || \qp(x_>|x_<)]}{control}
- \describe{\E[\lnqp(w|x)-\lnpp(w|x_<)]}{information gain}, \\
&\hspace*{-4em}
\describe{\E[\lnqp(w|x)-\lnpp(w|x_<)]}{information gain}
\geq \sum_{t'>t}
  \describe{\E[\lnqp(w|x_{1:t'})-\lnpp(w|x_{1:t'-1})]}{intrinsic reward}.
\label{eq:infogain}\raisetag{5ex}}

Information gain can be estimated by planning \citep{sun2011plansurprise} or from past environment interaction \citep{schmidhuber1991curiousmodel}. State representations lead to agents that disambiguate unobserved environment states, for example by opening doors to see objects behind them, such as in active inference \citep{da2020activeinference}, INDIGO \citep{azar2019ndigo}, and DVBF-LM \citep{mirchev2018dvbflm}. Model parameters lead to agents that discover the rules of their environment, such as
in active inference \citep{friston2015epistemics}, VIME \citep{houthooft2016vime}, MAX \citep{shyam2018max}, and Plan2Explore \citep{sekar2020plan2explore}. SLAM resolves uncertainty over both states and dynamics \citep{moutarlier1989slam}. Policy parameters lead to agents that explore to find the best behavior, such as bootstrapped DQN \citep{osband2016bootdqn} and Bayesian DQN \citep{azizzadenesheli2018bayesiandqn}.

One might think exploration should seek inputs with large error, but reconstruction and exploration optimize the same objective. Maximizing information gain minimizes the reconstruction error at future time steps by steering toward diverse but predictable inputs. Divergence minimization shows that every latent representation should be accompanied with an expected information gain term, so that the agent optimizes a consistent objective for past and future time steps. Moreover, it shows that representations should be optimized jointly with the policy to support both reconstruction and action choice \citep{lange2010rlrecon,jaderberg2016unreal,lee2019slac,yarats2019sacae}.

\section{Related Work}
\label{sec:related}

\paragraph{Divergence minimization}

Various problems have been formulated as minimizing a divergence between two distributions. TherML \citep{alemi2018therml} studies representation learning as KL minimization. We follow their interpretation of the data and belief as actual distribution, although their target is only defined by its factorization. ALICE \citep{li2017alice} describes adversarial learning as joint distribution matching, while \citet{kirsch2020ibframework} unify information-based objectives. \citet{ghasemipour2019divminimitation} describe imitation learning as minimizing divergences between the inputs of learned and expert behavior. None of these works consider combined representation learning and control. Thompson sampling minimizes the forward KL to explain action and perception as exact inference \citep{ortega2010bayesiancontrolrule}. In comparison, we optimize the backward KL to support intractable models and connect to a wide range of practical objectives.

\paragraph{Active inference}

The presented framework is inspired by the free energy principle, which studies the dynamics of agent and environment as stationary SDEs \citep{friston2010fep,friston2019physics}. We inherit the interpretations of active inference, which implements agents based on the free energy principle \citep{friston2017activeinference}. While divergence minimization matches the input distribution under the model, active inference maximizes the probability of inputs under it, resulting in smaller niches in heteroskedastic environments. Importantly, active inference optimizes the exploration terms only with respect to actions, which requires a specific action prior. Finally, typical implementations of active inference involve an expensive Bayesian model average over possible action sequences, limiting its applications to date \citep{friston2015epistemics,friston2020sophisticated}. We compare to active inference in detail in \cref{sec:actinf}. Generalized free energy \citep{parr2019gfe} studies a unified objective similar to ours, although its entropy terms are defined heuristically rather than derived from a general principle.

\paragraph{Control as inference}

It is well known that RL can be formulated as KL minimization over inputs and actions \citep{todorov2008duality,kappen2009klcontrol,rawlik2010controlasinference,ortega2011boundedrational,levine2018maxent}, as well as skills \citep{hausman2018skills,tirumala2019maxenttasks,galashov2019defaultpolicy}. We build upon this literature and extend it to agents with latent representations, leading to variational inference on past inputs and information seeking exploration for future inputs. Divergence minimization relates the above methods and motivates an additional entropy regularizer for inputs \citep{todorov2008duality,lee2019smm,xin2020explentropy}. SLAC \citep{lee2019slac} combines representation learning and control but does not consider the future mutual information, so their objective changes over time. In comparison, we derive the terms from a general principle and point out the information gain that results in an objective that is consistent over time. The information gain term may also address concerns about maximum entropy RL raised by \citet{odonoghue2020maxentsense}.

\section{Conclusion}
\label{sec:conclusion}

We introduce a general objective for action and perception of intelligent agents, based on minimizing the KL divergence. To unify the two processes, we formulate them as joint KL minimization with a shared target distribution. This target distribution is the probabilistic model under which the agent infers its representations and expresses the agent's preferences over system configurations. We summarize the key takeaways as follows:

\begin{itemize}
\itempar{Unified objective for action and perception} Divergence minimization with an expressive target maximizes the mutual information between latents and inputs. This leads to inferring representations that are informative of past inputs and exploration of future inputs that are informative of the representations. To optimize a consistent objective that does not change over time, any latent representation should be accompanied by a corresponding exploration term.
\itempar{Understanding of latent variables for decision making} Different latents lead to different objective terms. Latent representations are never observed, leading to both representation learning and information gain exploration. Actions and skills become observed over time and thus do not encourage representation learning but lead to generalized empowerment and skill discovery.
\itempar{Adaptive agents through expressive world models} Divergence minimization agents with an expressive target find niches where they can accurately predict their inputs and that they can inhabit despite external perturbations. The niches correspond to the inputs that the agent can learn to understand, which is facilitated by the exploration terms. This suggests designing powerful world models as a path toward building autonomous agents, without the need for task rewards.
\itempar{General recipe for designing novel objectives} When introducing new agent objectives, we recommend deriving them from the joint KL by choosing a latent structure and target. For information maximizing agents, the target is an expressive model, leaving different latent structures to be explored. Deriving novel objectives from the joint KL facilitates comparison, renders explicit the target distribution, and highlights the intrinsic objective terms needed to reach that distribution.
\itempar{Scalable objective for active inference} We connect the deep reinforcement learning and active inference literature. The joint KL is simpler and overcomes scalability limitations of typical active inference implementations, as evidenced by the many practically successful methods discussed in this paper. At the same time, the joint KL with latent representations contributes the expected information gain to the typical deep reinforcement learning objectives.
\itempar{Discovering new families of agent objectives} Our work shows that a family of representation learning and exploration objectives can be derived from minimizing a joint KL between the system and a target distribution. Different divergence measures give rise to new families of such agent objectives that could be easier to optimize or converge to better optima for infeasible targets. We leave exploring those objective families and comparing them empirically as future work.
\end{itemize}

Without constraining the class of targets, our framework is general and can describe any system. This by itself offers a framework for comparing many existing methods. However, interpreting the target as a model further suggests that intelligent agents may use especially expressive models as targets. This hypothesis should be investigated in future work by examining artificial agents with expressive world models or by modeling the behavior of natural agents as divergence minimization.

\paragraph{Acknowledgements}

Hidden for review.

We thank
Alex Alemi,
Julius Kunze,
Oleh Rybkin,
Kory Mathewson,
George Tucker,
Ian Fischer,
Simon Kornblith,
Ben Eysenbach,
Raza Habib,
Alex Immer,
Shane Gu,
Martin Biehl,
Adam Kosiorek,
Ben Poole,
Geoffrey Hinton,
Chen Li,
Louis Kirsch,
Dinesh Jayaraman,
Ricky Chen,
and James Davidson
for helpful discussions and feedback.

\clearpage
\begin{hyphenrules}{nohyphenation}
\bibliography{references}

\begin{thebibliography}{153}
\providecommand{\natexlab}[1]{#1}
\providecommand{\url}[1]{\texttt{#1}}
\expandafter\ifx\csname urlstyle\endcsname\relax
  \providecommand{\doi}[1]{doi: #1}\else
  \providecommand{\doi}{doi: \begingroup \urlstyle{rm}\Url}\fi

\bibitem[Achiam et~al.(2018)Achiam, Edwards, Amodei, and
  Abbeel]{achiam2018valor}
J~Achiam, H~Edwards, D~Amodei, P~Abbeel.
\newblock \titlecap{Variational option discovery algorithms}.
\newblock \emph{\titlecap{arXiv preprint arXiv:1807.10299}}, 2018.

\bibitem[Alemi and Fischer(2018)]{alemi2018therml}
AA~Alemi I~Fischer.
\newblock \titlecap{Ther{ML}: Thermodynamics of machine learning}.
\newblock \emph{\titlecap{arXiv preprint arXiv:1807.04162}}, 2018.

\bibitem[Amari(1967)]{amari1967onlinelearning}
S~Amari.
\newblock \titlecap{A theory of adaptive pattern classifiers}.
\newblock \emph{\titlecap{IEEE Transactions on Electronic Computers}}, 1967.

\bibitem[Ao et~al.(2013)Ao, Tian-Qi, and Jiang-Hong]{ao2013dynamicdecomp}
P~Ao, C~Tian-Qi, S~Jiang-Hong.
\newblock \titlecap{Dynamical decomposition of markov processes without
  detailed balance}.
\newblock \emph{\titlecap{Chinese Physics Letters}}, 30\penalty0 (7), 2013.

\bibitem[Ashby(1961)]{ashby1961cybernetics}
WR~Ashby.
\newblock \emph{\titlecap{An introduction to cybernetics}}.
\newblock Chapman \& Hall Ltd, 1961.

\bibitem[Azar et~al.(2019)Azar, Piot, Pires, Grill, Altch{\'e}, and
  Munos]{azar2019ndigo}
MG~Azar, B~Piot, BA~Pires, JB~Grill, F~Altch{\'e}, R~Munos.
\newblock \titlecap{World discovery models}.
\newblock \emph{\titlecap{arXiv preprint arXiv:1902.07685}}, 2019.

\bibitem[Azizzadenesheli et~al.(2018)Azizzadenesheli, Brunskill, and
  Anandkumar]{azizzadenesheli2018bayesiandqn}
K~Azizzadenesheli, E~Brunskill, A~Anandkumar.
\newblock \titlecap{Efficient exploration through bayesian deep
  {Q}-{N}etworks}.
\newblock \emph{\titlecap{2018 Information Theory and Applications Workshop
  (ITA)}}. IEEE, 2018.

\bibitem[Barber and Agakov(2003)]{barber2003variationalinfo}
D~Barber FV~Agakov.
\newblock \titlecap{The {IM} algorithm: a variational approach to information
  maximization}.
\newblock \emph{\titlecap{Advances in neural information processing systems}},
  2003.

\bibitem[Bellemare et~al.(2016)Bellemare, Srinivasan, Ostrovski, Schaul,
  Saxton, and Munos]{bellemare2016cts}
M~Bellemare, S~Srinivasan, G~Ostrovski, T~Schaul, D~Saxton, R~Munos.
\newblock \titlecap{Unifying count-based exploration and intrinsic motivation}.
\newblock \emph{\titlecap{Advances in Neural Information Processing Systems}},
  2016.

\bibitem[Berseth et~al.(2019)Berseth, Geng, Devin, Finn, Jayaraman, and
  Levine]{berseth2019smirl}
G~Berseth, D~Geng, C~Devin, C~Finn, D~Jayaraman, S~Levine.
\newblock \titlecap{Smirl: Surprise minimizing rl in dynamic environments}.
\newblock \emph{\titlecap{arXiv preprint arXiv:1912.05510}}, 2019.

\bibitem[Bishop(2006)]{bishop2006book}
CM~Bishop.
\newblock \emph{\titlecap{Pattern recognition and machine learning}}.
\newblock springer, 2006.

\bibitem[Blundell et~al.(2015)Blundell, Cornebise, Kavukcuoglu, and
  Wierstra]{blundell2015bbb}
C~Blundell, J~Cornebise, K~Kavukcuoglu, D~Wierstra.
\newblock \titlecap{Weight uncertainty in neural networks}.
\newblock \emph{\titlecap{arXiv preprint arXiv:1505.05424}}, 2015.

\bibitem[Bowling and Veloso(2000)]{bowling2000stochasticgames}
M~Bowling M~Veloso.
\newblock \titlecap{An analysis of stochastic game theory for multiagent
  reinforcement learning}.
\newblock Technical report, Carnegie-Mellon Univ Pittsburgh Pa School of
  Computer Science, 2000.

\bibitem[Bowman et~al.(2015)Bowman, Vilnis, Vinyals, Dai, Jozefowicz, and
  Bengio]{bowman2015klannealing}
SR~Bowman, L~Vilnis, O~Vinyals, AM~Dai, R~Jozefowicz, S~Bengio.
\newblock \titlecap{Generating sentences from a continuous space}.
\newblock \emph{\titlecap{arXiv preprint arXiv:1511.06349}}, 2015.

\bibitem[Brown(1981)]{brown1981completeclass}
LD~Brown.
\newblock \titlecap{A complete class theorem for statistical problems with
  finite sample spaces}.
\newblock \emph{\titlecap{The Annals of Statistics}}, 1981.

\bibitem[Burda et~al.(2018)Burda, Edwards, Storkey, and Klimov]{burda2018rnd}
Y~Burda, H~Edwards, A~Storkey, O~Klimov.
\newblock \titlecap{Exploration by random network distillation}.
\newblock \emph{\titlecap{arXiv preprint arXiv:1810.12894}}, 2018.

\bibitem[Chang et~al.(2019)Chang, Yao, Williams-King, and
  Lipson]{chang2019bnnensemble}
O~Chang, Y~Yao, D~Williams-King, H~Lipson.
\newblock \titlecap{Ensemble model patching: A parameter-efficient variational
  bayesian neural network}.
\newblock \emph{\titlecap{arXiv preprint arXiv:1905.09453}}, 2019.

\bibitem[Chen et~al.(2020)Chen, Kornblith, Norouzi, and Hinton]{chen2020simclr}
T~Chen, S~Kornblith, M~Norouzi, G~Hinton.
\newblock \titlecap{A simple framework for contrastive learning of visual
  representations}.
\newblock \emph{\titlecap{arXiv preprint arXiv:2002.05709}}, 2020.

\bibitem[Csisz{\'a}r and Matus(2003)]{csiszar2003projection}
I~Csisz{\'a}r F~Matus.
\newblock \titlecap{Information projections revisited}.
\newblock \emph{\titlecap{IEEE Transactions on Information Theory}},
  49\penalty0 (6), 2003.

\bibitem[Da~Costa et~al.(2020)Da~Costa, Parr, Sajid, Veselic, Neacsu, and
  Friston]{da2020activeinference}
L~Da~Costa, T~Parr, N~Sajid, S~Veselic, V~Neacsu, K~Friston.
\newblock \titlecap{Active inference on discrete state-spaces: a synthesis}.
\newblock \emph{\titlecap{arXiv preprint arXiv:2001.07203}}, 2020.

\bibitem[Dayan et~al.(1995)Dayan, Hinton, Neal, and Zemel]{dayan1995helmholtz}
P~Dayan, GE~Hinton, RM~Neal, RS~Zemel.
\newblock \titlecap{The {H}elmholtz machine}.
\newblock \emph{\titlecap{Neural computation}}, 7\penalty0 (5), 1995.

\bibitem[de~Abril and Kanai(2018)]{de2018curiosityempow}
IM~de~Abril R~Kanai.
\newblock \titlecap{A unified strategy for implementing curiosity and
  empowerment driven reinforcement learning}.
\newblock \emph{\titlecap{arXiv preprint arXiv:1806.06505}}, 2018.

\bibitem[Denker et~al.(1987)Denker, Schwartz, Wittner, Solla, Howard, Jackel,
  and Hopfield]{denker1987bnn}
J~Denker, D~Schwartz, B~Wittner, S~Solla, R~Howard, L~Jackel, J~Hopfield.
\newblock \titlecap{Large automatic learning, rule extraction, and
  generalization}.
\newblock \emph{\titlecap{Complex Systems}}, 1\penalty0 (5), 1987.

\bibitem[Dirac(1958)]{dirac1958quantummechanics}
PAM Dirac.
\newblock \emph{\titlecap{The principles of quantum mechanics}}.
\newblock Oxford university press, 1958.

\bibitem[Dusenberry et~al.(2020)Dusenberry, Jerfel, Wen, Ma, Snoek, Heller,
  Lakshminarayanan, and Tran]{dusenberry2020bnnrank1}
MW~Dusenberry, G~Jerfel, Y~Wen, Ya~Ma, J~Snoek, K~Heller, B~Lakshminarayanan,
  D~Tran.
\newblock \titlecap{Efficient and scalable bayesian neural nets with rank-1
  factors}.
\newblock \emph{\titlecap{arXiv preprint arXiv:2005.07186}}, 2020.

\bibitem[Ebert et~al.(2017)Ebert, Finn, Lee, and Levine]{ebert2017visualmpc}
F~Ebert, C~Finn, AX~Lee, S~Levine.
\newblock \titlecap{Self-supervised visual planning with temporal skip
  connections}.
\newblock \emph{\titlecap{arXiv preprint arXiv:1710.05268}}, 2017.

\bibitem[Ebert et~al.(2018)Ebert, Finn, Dasari, Xie, Lee, and
  Levine]{ebert2018foresight}
F~Ebert, C~Finn, S~Dasari, A~Xie, A~Lee, S~Levine.
\newblock \titlecap{Visual foresight: Model-based deep reinforcement learning
  for vision-based robotic control}.
\newblock \emph{\titlecap{arXiv preprint arXiv:1812.00568}}, 2018.

\bibitem[Eslami et~al.(2018)Eslami, Rezende, Besse, Viola, Morcos, Garnelo,
  Ruderman, Rusu, Danihelka, Gregor, et~al.]{eslami2018gqn}
SA~Eslami, DJ~Rezende, F~Besse, F~Viola, AS~Morcos, M~Garnelo, A~Ruderman,
  AA~Rusu, I~Danihelka, K~Gregor, et~al.
\newblock \titlecap{Neural scene representation and rendering}.
\newblock \emph{\titlecap{Science}}, 360\penalty0 (6394), 2018.

\bibitem[Eysenbach et~al.(2018)Eysenbach, Gupta, Ibarz, and
  Levine]{eysenbach2018diayn}
B~Eysenbach, A~Gupta, J~Ibarz, S~Levine.
\newblock \titlecap{Diversity is all you need: learning skills without a reward
  function}.
\newblock \emph{\titlecap{arXiv preprint arXiv:1802.06070}}, 2018.

\bibitem[Fischer(2020)]{fischer2020ceb}
I~Fischer.
\newblock \titlecap{The conditional entropy bottleneck}.
\newblock \emph{\titlecap{arXiv preprint arXiv:2002.05379}}, 2020.

\bibitem[Florensa et~al.(2017)Florensa, Duan, and Abbeel]{florensa2017snn}
C~Florensa, Y~Duan, P~Abbeel.
\newblock \titlecap{Stochastic neural networks for hierarchical reinforcement
  learning}.
\newblock \emph{\titlecap{arXiv preprint arXiv:1704.03012}}, 2017.

\bibitem[Fox et~al.(2015)Fox, Pakman, and Tishby]{fox2015softq}
R~Fox, A~Pakman, N~Tishby.
\newblock \titlecap{Taming the noise in reinforcement learning via soft
  updates}.
\newblock \emph{\titlecap{arXiv preprint arXiv:1512.08562}}, 2015.

\bibitem[Friston(2010)]{friston2010fep}
K~Friston.
\newblock \titlecap{The free-energy principle: a unified brain theory?}
\newblock \emph{\titlecap{Nature reviews neuroscience}}, 11\penalty0 (2), 2010.

\bibitem[Friston(2013)]{friston2013life}
K~Friston.
\newblock \titlecap{Life as we know it}.
\newblock \emph{\titlecap{Journal of the Royal Society Interface}}, 10\penalty0
  (86), 2013.

\bibitem[Friston(2019)]{friston2019physics}
K~Friston.
\newblock \titlecap{A free energy principle for a particular physics}.
\newblock \emph{\titlecap{arXiv preprint arXiv:1906.10184}}, 2019.

\bibitem[Friston et~al.(2012)Friston, Adams, and Montague]{friston2012value}
K~Friston, R~Adams, R~Montague.
\newblock \titlecap{What is value--accumulated reward or evidence?}
\newblock \emph{\titlecap{Frontiers in neurorobotics}}, 6, 2012.

\bibitem[Friston et~al.(2015)Friston, Rigoli, Ognibene, Mathys, Fitzgerald, and
  Pezzulo]{friston2015epistemics}
K~Friston, F~Rigoli, D~Ognibene, C~Mathys, T~Fitzgerald, G~Pezzulo.
\newblock \titlecap{Active inference and epistemic value}.
\newblock \emph{\titlecap{Cognitive neuroscience}}, 6\penalty0 (4), 2015.

\bibitem[Friston et~al.(2017)Friston, FitzGerald, Rigoli, Schwartenbeck, and
  Pezzulo]{friston2017activeinference}
K~Friston, T~FitzGerald, F~Rigoli, P~Schwartenbeck, G~Pezzulo.
\newblock \titlecap{Active inference: a process theory}.
\newblock \emph{\titlecap{Neural computation}}, 29\penalty0 (1), 2017.

\bibitem[Friston et~al.(2020)Friston, Da~Costa, Hafner, Hesp, and
  Parr]{friston2020sophisticated}
K~Friston, L~Da~Costa, D~Hafner, C~Hesp, T~Parr.
\newblock \titlecap{Sophisticated inference}.
\newblock \emph{\titlecap{arXiv preprint arXiv:2006.04120}}, 2020.

\bibitem[Galashov et~al.(2019)Galashov, Jayakumar, Hasenclever, Tirumala,
  Schwarz, Desjardins, Czarnecki, Teh, Pascanu, and
  Heess]{galashov2019defaultpolicy}
A~Galashov, SM~Jayakumar, L~Hasenclever, D~Tirumala, J~Schwarz, G~Desjardins,
  WM~Czarnecki, YW~Teh, R~Pascanu, N~Heess.
\newblock \titlecap{Information asymmetry in kl-regularized rl}.
\newblock \emph{\titlecap{arXiv preprint arXiv:1905.01240}}, 2019.

\bibitem[Ghahramani and Jordan(1995)]{ghahramani1995missing}
Z~Ghahramani MI~Jordan.
\newblock \titlecap{Learning from incomplete data}, 1995.

\bibitem[Ghasemipour et~al.(2019)Ghasemipour, Zemel, and
  Gu]{ghasemipour2019divminimitation}
SKS Ghasemipour, R~Zemel, S~Gu.
\newblock \titlecap{A divergence minimization perspective on imitation learning
  methods}.
\newblock \emph{\titlecap{arXiv preprint arXiv:1911.02256}}, 2019.

\bibitem[Ghosh and Doshi-Velez(2017)]{ghosh2017bnnhorseshoe}
S~Ghosh F~Doshi-Velez.
\newblock \titlecap{Model selection in bayesian neural networks via horseshoe
  priors}.
\newblock \emph{\titlecap{arXiv preprint arXiv:1705.10388}}, 2017.

\bibitem[Gregor et~al.(2016)Gregor, Rezende, and Wierstra]{gregor2016vic}
K~Gregor, DJ~Rezende, D~Wierstra.
\newblock \titlecap{Variational intrinsic control}.
\newblock \emph{\titlecap{arXiv preprint arXiv:1611.07507}}, 2016.

\bibitem[Gregor et~al.(2019)Gregor, Rezende, Besse, Wu, Merzic, and
  Oord]{gregor2019beliefstate}
K~Gregor, DJ~Rezende, F~Besse, Y~Wu, H~Merzic, Avd Oord.
\newblock \titlecap{Shaping belief states with generative environment models
  for rl}.
\newblock \emph{\titlecap{arXiv preprint arXiv:1906.09237}}, 2019.

\bibitem[Gregory(1980)]{gregory1980perception}
RL~Gregory.
\newblock \titlecap{Perceptions as hypotheses}.
\newblock \emph{\titlecap{Philosophical Transactions of the Royal Society of
  London. B, Biological Sciences}}, 290\penalty0 (1038), 1980.

\bibitem[Guo et~al.(2018)Guo, Azar, Piot, Pires, Pohlen, and
  Munos]{guo2018actioncpc}
ZD~Guo, MG~Azar, B~Piot, BA~Pires, T~Pohlen, R~Munos.
\newblock \titlecap{Neural predictive belief representations}.
\newblock \emph{\titlecap{arXiv preprint arXiv:1811.06407}}, 2018.

\bibitem[Gutmann and Hyv{\"a}rinen(2010)]{gutmann2010nce}
M~Gutmann A~Hyv{\"a}rinen.
\newblock \titlecap{Noise-contrastive estimation: a new estimation principle
  for unnormalized statistical models}.
\newblock \emph{\titlecap{Proceedings of the Thirteenth International
  Conference on Artificial Intelligence and Statistics}}, 2010.

\bibitem[Ha and Schmidhuber(2018)]{ha2018worldmodels}
D~Ha J~Schmidhuber.
\newblock \titlecap{World models}.
\newblock \emph{\titlecap{arXiv preprint arXiv:1803.10122}}, 2018.

\bibitem[Ha et~al.(2016)Ha, Dai, and Le]{ha2016hypernets}
D~Ha, A~Dai, QV~Le.
\newblock \titlecap{Hypernetworks}.
\newblock \emph{\titlecap{arXiv preprint arXiv:1609.09106}}, 2016.

\bibitem[Haarnoja et~al.(2017)Haarnoja, Tang, Abbeel, and
  Levine]{haarnoja2017sql}
T~Haarnoja, H~Tang, P~Abbeel, S~Levine.
\newblock \titlecap{Reinforcement learning with deep energy-based policies}.
\newblock \emph{\titlecap{Proceedings of the 34th International Conference on
  Machine Learning-Volume 70}}. JMLR. org, 2017.

\bibitem[Haarnoja et~al.(2018)Haarnoja, Zhou, Abbeel, and
  Levine]{haarnoja2018sac}
T~Haarnoja, A~Zhou, P~Abbeel, S~Levine.
\newblock \titlecap{Soft actor-critic: Off-policy maximum entropy deep
  reinforcement learning with a stochastic actor}.
\newblock \emph{\titlecap{arXiv preprint arXiv:1801.01290}}, 2018.

\bibitem[Hafner et~al.(2018)Hafner, Lillicrap, Fischer, Villegas, Ha, Lee, and
  Davidson]{hafner2018planet}
D~Hafner, T~Lillicrap, I~Fischer, R~Villegas, D~Ha, H~Lee, J~Davidson.
\newblock \titlecap{Learning latent dynamics for planning from pixels}.
\newblock \emph{\titlecap{arXiv preprint arXiv:1811.04551}}, 2018.

\bibitem[Hafner et~al.(2019{\natexlab{a}})Hafner, Lillicrap, Ba, and
  Norouzi]{hafner2019dreamer}
D~Hafner, T~Lillicrap, J~Ba, M~Norouzi.
\newblock \titlecap{Dream to control: Learning behaviors by latent
  imagination}.
\newblock \emph{\titlecap{arXiv preprint arXiv:1912.01603}},
  2019{\natexlab{a}}.

\bibitem[Hafner et~al.(2019{\natexlab{b}})Hafner, Tran, Irpan, Lillicrap, and
  Davidson]{hafner2018ncp}
D~Hafner, D~Tran, A~Irpan, T~Lillicrap, J~Davidson.
\newblock \titlecap{Reliable uncertainty estimates in deep neural networks
  using noise contrastive priors}.
\newblock \emph{\titlecap{Conference on Uncertainty in Artificial
  Intelligence}}, 2019{\natexlab{b}}.

\bibitem[Haken(1981)]{haken1981synergetics}
H~Haken.
\newblock \emph{\titlecap{The science of structure: Synergetics}}.
\newblock Van Nostrand Reinhold, 1981.

\bibitem[Hausman et~al.(2018)Hausman, Springenberg, Wang, Heess, and
  Riedmiller]{hausman2018skills}
K~Hausman, JT~Springenberg, Z~Wang, N~Heess, M~Riedmiller.
\newblock \titlecap{Learning an embedding space for transferable robot skills}.
\newblock \emph{\titlecap{International Conference on Learning
  Representations}}, 2018.

\bibitem[Helmholtz(1866)]{helmholtz1866perception}
Hv~Helmholtz.
\newblock \titlecap{Concerning the perceptions in general}.
\newblock \emph{\titlecap{Treatise on physiological optics}}, 1866.

\bibitem[Hinton et~al.(2015)Hinton, Vinyals, and Dean]{hinton2015distillation}
G~Hinton, O~Vinyals, J~Dean.
\newblock \titlecap{Distilling the knowledge in a neural network}.
\newblock \emph{\titlecap{arXiv preprint arXiv:1503.02531}}, 2015.

\bibitem[Hinton and Van~Camp(1993)]{hinton1993vi}
GE~Hinton D~Van~Camp.
\newblock \titlecap{Keeping the neural networks simple by minimizing the
  description length of the weights}.
\newblock \emph{\titlecap{Proceedings of the sixth annual conference on
  Computational learning theory}}, 1993.

\bibitem[Hinton et~al.(2006)Hinton, Osindero, and Teh]{hinton2006deepbelief}
GE~Hinton, S~Osindero, YW~Teh.
\newblock \titlecap{A fast learning algorithm for deep belief nets}.
\newblock \emph{\titlecap{Neural computation}}, 18\penalty0 (7), 2006.

\bibitem[Houthooft et~al.(2016)Houthooft, Chen, Duan, Schulman, De~Turck, and
  Abbeel]{houthooft2016vime}
R~Houthooft, X~Chen, Y~Duan, J~Schulman, F~De~Turck, P~Abbeel.
\newblock \titlecap{{VIME}: Variational information maximizing exploration}.
\newblock \emph{\titlecap{Advances in Neural Information Processing Systems}},
  2016.

\bibitem[Howard and Matheson(1972)]{howard1972riskrl}
RA~Howard JE~Matheson.
\newblock \titlecap{Risk-sensitive markov decision processes}.
\newblock \emph{\titlecap{Management science}}, 18\penalty0 (7), 1972.

\bibitem[Huffman(1952)]{huffman1952coding}
DA~Huffman.
\newblock \titlecap{A method for the construction of minimum-redundancy codes}.
\newblock \emph{\titlecap{Proceedings of the IRE}}, 40\penalty0 (9), 1952.

\bibitem[Immer et~al.(2020)Immer, Korzepa, and Bauer]{immer2020bnnlocal}
A~Immer, M~Korzepa, M~Bauer.
\newblock \titlecap{Improving predictions of bayesian neural networks via local
  linearization}.
\newblock \emph{\titlecap{arXiv preprint arXiv:2008.08400}}, 2020.

\bibitem[Izmailov et~al.(2018)Izmailov, Podoprikhin, Garipov, Vetrov, and
  Wilson]{izmailov2018bnnpca}
P~Izmailov, D~Podoprikhin, T~Garipov, D~Vetrov, AG~Wilson.
\newblock \titlecap{Averaging weights leads to wider optima and better
  generalization}.
\newblock \emph{\titlecap{arXiv preprint arXiv:1803.05407}}, 2018.

\bibitem[Jaderberg et~al.(2016)Jaderberg, Mnih, Czarnecki, Schaul, Leibo,
  Silver, and Kavukcuoglu]{jaderberg2016unreal}
M~Jaderberg, V~Mnih, WM~Czarnecki, T~Schaul, JZ~Leibo, D~Silver, K~Kavukcuoglu.
\newblock \titlecap{Reinforcement learning with unsupervised auxiliary tasks}.
\newblock \emph{\titlecap{arXiv preprint arXiv:1611.05397}}, 2016.

\bibitem[Jaynes(1957)]{jaynes1957maximumentropy}
ET~Jaynes.
\newblock \titlecap{Information theory and statistical mechanics}.
\newblock \emph{\titlecap{Physical review}}, 106\penalty0 (4), 1957.

\bibitem[Jefferys and Berger(1992)]{jefferys1992occamsrazor}
WH~Jefferys JO~Berger.
\newblock \titlecap{Ockham's razor and bayesian analysis}.
\newblock \emph{\titlecap{American Scientist}}, 80\penalty0 (1), 1992.

\bibitem[Jeffreys(1935)]{jeffreys1935bayesfactor}
H~Jeffreys.
\newblock \titlecap{Some tests of significance, treated by the theory of
  probability}.
\newblock \emph{\titlecap{Mathematical Proceedings of the Cambridge
  Philosophical Society}}, volume~31. Cambridge University Press, 1935.

\bibitem[Jordan et~al.(1999)Jordan, Ghahramani, Jaakkola, and
  Saul]{jordan1999vi}
MI~Jordan, Z~Ghahramani, TS~Jaakkola, LK~Saul.
\newblock \titlecap{An introduction to variational methods for graphical
  models}.
\newblock \emph{\titlecap{Machine learning}}, 37\penalty0 (2), 1999.

\bibitem[Kalman(1960)]{kalman1960filter}
RE~Kalman.
\newblock \titlecap{A new approach to linear filtering and prediction
  problems}.
\newblock \emph{\titlecap{Journal of basic Engineering}}, 82\penalty0 (1),
  1960.

\bibitem[Kappen et~al.(2009)Kappen, G{\'o}mez, and Opper]{kappen2009klcontrol}
HJ~Kappen, V~G{\'o}mez, M~Opper.
\newblock \titlecap{Optimal control as a graphical model inference problem}.
\newblock \emph{\titlecap{Machine learning}}, 87\penalty0 (2), 2009.

\bibitem[Karl et~al.(2016)Karl, Soelch, Bayer, and van~der Smagt]{karl2016dvbf}
M~Karl, M~Soelch, J~Bayer, P~van~der Smagt.
\newblock \titlecap{Deep variational bayes filters: Unsupervised learning of
  state space models from raw data}.
\newblock \emph{\titlecap{arXiv preprint arXiv:1605.06432}}, 2016.

\bibitem[Karl et~al.(2017)Karl, Soelch, Becker-Ehmck, Benbouzid, van~der Smagt,
  and Bayer]{karl2017empowerment}
M~Karl, M~Soelch, P~Becker-Ehmck, D~Benbouzid, P~van~der Smagt, J~Bayer.
\newblock \titlecap{Unsupervised real-time control through variational
  empowerment}.
\newblock \emph{\titlecap{arXiv preprint arXiv:1710.05101}}, 2017.

\bibitem[Kass and Raftery(1995)]{kass1995bayesfactor}
RE~Kass AE~Raftery.
\newblock \titlecap{Bayes factors}.
\newblock \emph{\titlecap{Journal of the american statistical association}},
  90\penalty0 (430), 1995.

\bibitem[Kim et~al.(2018)Kim, Wiseman, Miller, Sontag, and
  Rush]{kim2018semiamortized}
Y~Kim, S~Wiseman, AC~Miller, D~Sontag, AM~Rush.
\newblock \titlecap{Semi-amortized variational autoencoders}.
\newblock \emph{\titlecap{arXiv preprint arXiv:1802.02550}}, 2018.

\bibitem[Kingma and Welling(2013)]{kingma2013vae}
DP~Kingma M~Welling.
\newblock \titlecap{Auto-encoding variational bayes}.
\newblock \emph{\titlecap{arXiv preprint arXiv:1312.6114}}, 2013.

\bibitem[Kirsch et~al.(2020)Kirsch, Lyle, and Gal]{kirsch2020ibframework}
A~Kirsch, C~Lyle, Y~Gal.
\newblock \titlecap{Unpacking information bottlenecks: Unifying
  information-theoretic objectives in deep learning}.
\newblock \emph{\titlecap{arXiv preprint arXiv:2003.12537}}, 2020.

\bibitem[Klyubin et~al.(2005)Klyubin, Polani, and
  Nehaniv]{klyubin2005empowerment}
AS~Klyubin, D~Polani, CL~Nehaniv.
\newblock \titlecap{Empowerment: A universal agent-centric measure of control}.
\newblock \emph{\titlecap{IEEE Congress on Evolutionary Computation}},
  volume~1. IEEE, 2005.

\bibitem[Krishnan et~al.(2015)Krishnan, Shalit, and
  Sontag]{krishnan2015deepkalman}
RG~Krishnan, U~Shalit, D~Sontag.
\newblock \titlecap{Deep kalman filters}.
\newblock \emph{\titlecap{arXiv preprint arXiv:1511.05121}}, 2015.

\bibitem[Kullback and Leibler(1951)]{kullback1951kl}
S~Kullback RA~Leibler.
\newblock \titlecap{On information and sufficiency}.
\newblock \emph{\titlecap{The annals of mathematical statistics}}, 22\penalty0
  (1), 1951.

\bibitem[Lange and Riedmiller(2010)]{lange2010rlrecon}
S~Lange M~Riedmiller.
\newblock \titlecap{Deep auto-encoder neural networks in reinforcement
  learning}.
\newblock \emph{\titlecap{The 2010 International Joint Conference on Neural
  Networks (IJCNN)}}. IEEE, 2010.

\bibitem[LeCun et~al.(2006)LeCun, Chopra, Hadsell, Ranzato, and
  Huang]{lecun2006energytutorial}
Y~LeCun, S~Chopra, R~Hadsell, M~Ranzato, F~Huang.
\newblock \titlecap{A tutorial on energy-based learning}.
\newblock \emph{\titlecap{Predicting structured data}}, 1\penalty0 (0), 2006.

\bibitem[Lee et~al.(2019{\natexlab{a}})Lee, Nagabandi, Abbeel, and
  Levine]{lee2019slac}
AX~Lee, A~Nagabandi, P~Abbeel, S~Levine.
\newblock \titlecap{Stochastic latent actor-critic: Deep reinforcement learning
  with a latent variable model}.
\newblock \emph{\titlecap{arXiv preprint arXiv:1907.00953}},
  2019{\natexlab{a}}.

\bibitem[Lee et~al.(2019{\natexlab{b}})Lee, Eysenbach, Parisotto, Xing, Levine,
  and Salakhutdinov]{lee2019smm}
L~Lee, B~Eysenbach, E~Parisotto, E~Xing, S~Levine, R~Salakhutdinov.
\newblock \titlecap{Efficient exploration via state marginal matching}.
\newblock \emph{\titlecap{arXiv preprint arXiv:1906.05274}},
  2019{\natexlab{b}}.

\bibitem[Leibfried et~al.(2019)Leibfried, Pascual-Diaz, and
  Grau-Moya]{leibfried2019empowreward}
F~Leibfried, S~Pascual-Diaz, J~Grau-Moya.
\newblock \titlecap{A unified bellman optimality principle combining reward
  maximization and empowerment}.
\newblock \emph{\titlecap{Advances in Neural Information Processing Systems}},
  2019.

\bibitem[Levine(2018)]{levine2018maxent}
S~Levine.
\newblock \titlecap{Reinforcement learning and control as probabilistic
  inference: Tutorial and review}.
\newblock \emph{\titlecap{arXiv preprint arXiv:1805.00909}}, 2018.

\bibitem[Li et~al.(2017)Li, Liu, Chen, Pu, Chen, Henao, and Carin]{li2017alice}
C~Li, H~Liu, C~Chen, Y~Pu, L~Chen, R~Henao, L~Carin.
\newblock \titlecap{Alice: Towards understanding adversarial learning for joint
  distribution matching}.
\newblock \emph{\titlecap{Advances in Neural Information Processing Systems}},
  2017.

\bibitem[Lindley et~al.(1956)]{lindley1956expectedinfo}
DV~Lindley et~al.
\newblock \titlecap{On a measure of the information provided by an experiment}.
\newblock \emph{\titlecap{The Annals of Mathematical Statistics}}, 27\penalty0
  (4), 1956.

\bibitem[Louizos and Welling(2016)]{louizos2016bnnmatrix}
C~Louizos M~Welling.
\newblock \titlecap{Structured and efficient variational deep learning with
  matrix gaussian posteriors}.
\newblock \emph{\titlecap{International Conference on Machine Learning}}, 2016.

\bibitem[Louizos and Welling(2017)]{louizos2017bnnflow}
C~Louizos M~Welling.
\newblock \titlecap{Multiplicative normalizing flows for variational bayesian
  neural networks}.
\newblock \emph{\titlecap{arXiv preprint arXiv:1703.01961}}, 2017.

\bibitem[Ma et~al.(2015)Ma, Chen, and Fox]{ma2015mcmcdecomp}
YA~Ma, T~Chen, E~Fox.
\newblock \titlecap{A complete recipe for stochastic gradient mcmc}.
\newblock \emph{\titlecap{Advances in Neural Information Processing Systems}},
  2015.

\bibitem[MacKay(1992{\natexlab{a}})]{mackay1992bnn}
DJ~MacKay.
\newblock \titlecap{A practical bayesian framework for backpropagation
  networks}.
\newblock \emph{\titlecap{Neural computation}}, 4\penalty0 (3),
  1992{\natexlab{a}}.

\bibitem[MacKay(1992{\natexlab{b}})]{mackay1992infogain}
DJ~MacKay.
\newblock \titlecap{Information-based objective functions for active data
  selection}.
\newblock \emph{\titlecap{Neural computation}}, 4\penalty0 (4),
  1992{\natexlab{b}}.

\bibitem[MacKay(2003)]{mackay2003information}
DJ~MacKay.
\newblock \emph{\titlecap{Information theory, inference and learning
  algorithms}}.
\newblock Cambridge university press, 2003.

\bibitem[Mirchev et~al.(2018)Mirchev, Kayalibay, Soelch, van~der Smagt, and
  Bayer]{mirchev2018dvbflm}
A~Mirchev, B~Kayalibay, M~Soelch, P~van~der Smagt, J~Bayer.
\newblock \titlecap{Approximate bayesian inference in spatial environments}.
\newblock \emph{\titlecap{arXiv preprint arXiv:1805.07206}}, 2018.

\bibitem[Mnih et~al.(2016)Mnih, Badia, Mirza, Graves, Lillicrap, Harley,
  Silver, and Kavukcuoglu]{mnih2016a3c}
V~Mnih, AP~Badia, M~Mirza, A~Graves, T~Lillicrap, T~Harley, D~Silver,
  K~Kavukcuoglu.
\newblock \titlecap{Asynchronous methods for deep reinforcement learning}.
\newblock \emph{\titlecap{International Conference on Machine Learning}}, 2016.

\bibitem[Mohamed and Rezende(2015)]{mohamed2015empowerment}
S~Mohamed DJ~Rezende.
\newblock \titlecap{Variational information maximisation for intrinsically
  motivated reinforcement learning}.
\newblock \emph{\titlecap{Advances in neural information processing systems}},
  2015.

\bibitem[Morgenstern and Von~Neumann(1953)]{morgenstern1953seu}
O~Morgenstern J~Von~Neumann.
\newblock \emph{\titlecap{Theory of games and economic behavior}}.
\newblock Princeton university press, 1953.

\bibitem[Moutarlier and Chatila(1989)]{moutarlier1989slam}
P~Moutarlier R~Chatila.
\newblock \titlecap{Stochastic multisensory data fusion for mobile robot
  location and environment modelling. 5th int}.
\newblock \emph{\titlecap{Symposium on Robotics Research}}, 1989.

\bibitem[Murphy(2012)]{murphy2012book}
KP~Murphy.
\newblock \emph{\titlecap{Machine learning: a probabilistic perspective}}.
\newblock MIT press, 2012.

\bibitem[O'Donoghue et~al.(2020)O'Donoghue, Osband, and
  Ionescu]{odonoghue2020maxentsense}
B~O'Donoghue, I~Osband, C~Ionescu.
\newblock \titlecap{Making sense of reinforcement learning and probabilistic
  inference}.
\newblock \emph{\titlecap{arXiv preprint arXiv:2001.00805}}, 2020.

\bibitem[Okada et~al.(2020)Okada, Kosaka, and
  Taniguchi]{okada2020bayesianplanet}
M~Okada, N~Kosaka, T~Taniguchi.
\newblock \titlecap{Planet of the bayesians: Reconsidering and improving deep
  planning network by incorporating bayesian inference}.
\newblock \emph{\titlecap{arXiv preprint arXiv:2003.00370}}, 2020.

\bibitem[Oord et~al.(2018)Oord, Li, and Vinyals]{oord2018cpc}
Avd Oord, Y~Li, O~Vinyals.
\newblock \titlecap{Representation learning with contrastive predictive
  coding}.
\newblock \emph{\titlecap{arXiv preprint arXiv:1807.03748}}, 2018.

\bibitem[Ortega and Braun(2011)]{ortega2011boundedrational}
DA~Ortega PA~Braun.
\newblock \titlecap{Information, utility and bounded rationality}.
\newblock \emph{\titlecap{International Conference on Artificial General
  Intelligence}}. Springer, 2011.

\bibitem[Ortega and Braun(2010)]{ortega2010bayesiancontrolrule}
PA~Ortega DA~Braun.
\newblock \titlecap{A minimum relative entropy principle for learning and
  acting}.
\newblock \emph{\titlecap{Journal of Artificial Intelligence Research}}, 38,
  2010.

\bibitem[Osband et~al.(2016)Osband, Blundell, Pritzel, and
  Van~Roy]{osband2016bootdqn}
I~Osband, C~Blundell, A~Pritzel, B~Van~Roy.
\newblock \titlecap{Deep exploration via bootstrapped {DQN}}.
\newblock \emph{\titlecap{Advances in neural information processing systems}},
  2016.

\bibitem[Oudeyer et~al.(2007)Oudeyer, Kaplan, and Hafner]{oudeyer2007curiosity}
PY~Oudeyer, F~Kaplan, VV~Hafner.
\newblock \titlecap{Intrinsic motivation systems for autonomous mental
  development}.
\newblock \emph{\titlecap{IEEE transactions on evolutionary computation}},
  11\penalty0 (2), 2007.

\bibitem[Parr and Friston(2019)]{parr2019gfe}
T~Parr KJ~Friston.
\newblock \titlecap{Generalised free energy and active inference}.
\newblock \emph{\titlecap{Biological cybernetics}}, 113\penalty0 (5-6), 2019.

\bibitem[Pathak et~al.(2017)Pathak, Agrawal, Efros, and Darrell]{pathak2017icm}
D~Pathak, P~Agrawal, AA~Efros, T~Darrell.
\newblock \titlecap{Curiosity-driven exploration by self-supervised
  prediction}.
\newblock \emph{\titlecap{Proceedings of the IEEE Conference on Computer Vision
  and Pattern Recognition Workshops}}, 2017.

\bibitem[Pearl(1995)]{pearl1995docalculus}
J~Pearl.
\newblock \titlecap{Causal diagrams for empirical research}.
\newblock \emph{\titlecap{Biometrika}}, 82\penalty0 (4), 1995.

\bibitem[Peterson(1987)]{peterson1987vi}
C~Peterson.
\newblock \titlecap{A mean field theory learning algorithm for neural
  networks}.
\newblock \emph{\titlecap{Complex systems}}, 1, 1987.

\bibitem[Pong et~al.(2019)Pong, Dalal, Lin, Nair, Bahl, and
  Levine]{pong2019skewfit}
VH~Pong, M~Dalal, S~Lin, A~Nair, S~Bahl, S~Levine.
\newblock \titlecap{Skew-fit: State-covering self-supervised reinforcement
  learning}.
\newblock \emph{\titlecap{arXiv preprint arXiv:1903.03698}}, 2019.

\bibitem[Poole et~al.(2019)Poole, Ozair, Oord, Alemi, and
  Tucker]{poole2019bounds}
B~Poole, S~Ozair, Avd Oord, AA~Alemi, G~Tucker.
\newblock \titlecap{On variational bounds of mutual information}.
\newblock \emph{\titlecap{arXiv preprint arXiv:1905.06922}}, 2019.

\bibitem[Pratt(1964)]{pratt1964risk}
JW~Pratt.
\newblock \titlecap{Risk aversion in the small and in the large}.
\newblock \emph{\titlecap{Econometrica}}, 32\penalty0 (1/2), 1964.

\bibitem[Rawlik et~al.(2010)Rawlik, Toussaint, and
  Vijayakumar]{rawlik2010controlasinference}
K~Rawlik, M~Toussaint, S~Vijayakumar.
\newblock \titlecap{Approximate inference and stochastic optimal control}.
\newblock \emph{\titlecap{arXiv preprint arXiv:1009.3958}}, 2010.

\bibitem[Rezende et~al.(2014)Rezende, Mohamed, and Wierstra]{rezende2014vae}
DJ~Rezende, S~Mohamed, D~Wierstra.
\newblock \titlecap{Stochastic backpropagation and approximate inference in
  deep generative models}.
\newblock \emph{\titlecap{arXiv preprint arXiv:1401.4082}}, 2014.

\bibitem[Rudin(1966)]{rudin1966analysis}
R~Rudin.
\newblock \titlecap{Complex analysis}, 1966.

\bibitem[Salge et~al.(2013)Salge, Glackin, and
  Polani]{salge2013continuousempow}
C~Salge, C~Glackin, D~Polani.
\newblock \titlecap{Approximation of empowerment in the continuous domain}.
\newblock \emph{\titlecap{Advances in Complex Systems}}, 16\penalty0 (02n03),
  2013.

\bibitem[Salge et~al.(2014)Salge, Glackin, and Polani]{salge2014realizedempow}
C~Salge, C~Glackin, D~Polani.
\newblock \titlecap{Changing the environment based on empowerment as intrinsic
  motivation}.
\newblock \emph{\titlecap{Entropy}}, 16\penalty0 (5), 2014.

\bibitem[Savinov et~al.(2018)Savinov, Raichuk, Marinier, Vincent, Pollefeys,
  Lillicrap, and Gelly]{savinov2018reachability}
N~Savinov, A~Raichuk, R~Marinier, D~Vincent, M~Pollefeys, T~Lillicrap, S~Gelly.
\newblock \titlecap{Episodic curiosity through reachability}.
\newblock \emph{\titlecap{arXiv preprint arXiv:1810.02274}}, 2018.

\bibitem[Schmidhuber(1991)]{schmidhuber1991curiousmodel}
J~Schmidhuber.
\newblock \titlecap{Curious model-building control systems}.
\newblock \emph{\titlecap{[Proceedings] 1991 IEEE International Joint
  Conference on Neural Networks}}. IEEE, 1991.

\bibitem[Schr{\"o}dinger(1944)]{schrodinger1944life}
E~Schr{\"o}dinger.
\newblock \emph{\titlecap{What is life? The physical aspect of the living cell
  and mind}}.
\newblock Cambridge University Press Cambridge, 1944.

\bibitem[Schulman et~al.(2017)Schulman, Chen, and
  Abbeel]{schulman2017equivalence}
J~Schulman, X~Chen, P~Abbeel.
\newblock \titlecap{Equivalence between policy gradients and soft q-learning}.
\newblock \emph{\titlecap{arXiv preprint arXiv:1704.06440}}, 2017.

\bibitem[Sekar et~al.(2020)Sekar, Rybkin, Daniilidis, Abbeel, Hafner, and
  Pathak]{sekar2020plan2explore}
R~Sekar, O~Rybkin, K~Daniilidis, P~Abbeel, D~Hafner, D~Pathak.
\newblock \titlecap{Planning to explore via self-supervised world models}.
\newblock \emph{\titlecap{arXiv preprint arXiv:2005.05960}}, 2020.

\bibitem[Shankar and Gupta(2020)]{shankar2020temporalvi}
T~Shankar A~Gupta.
\newblock \titlecap{Learning robot skills with temporal variational inference}.
\newblock \emph{\titlecap{arXiv preprint arXiv:2006.16232}}, 2020.

\bibitem[Shannon(1948)]{shannon1948infotheory}
CE~Shannon.
\newblock \titlecap{A mathematical theory of communication}.
\newblock \emph{\titlecap{Bell system technical journal}}, 27\penalty0 (3),
  1948.

\bibitem[Sharma et~al.(2019)Sharma, Gu, Levine, Kumar, and
  Hausman]{sharma2019dads}
A~Sharma, S~Gu, S~Levine, V~Kumar, K~Hausman.
\newblock \titlecap{Dynamics-aware unsupervised discovery of skills}.
\newblock \emph{\titlecap{arXiv preprint arXiv:1907.01657}}, 2019.

\bibitem[Shyam et~al.(2018)Shyam, Ja{\'s}kowski, and Gomez]{shyam2018max}
P~Shyam, W~Ja{\'s}kowski, F~Gomez.
\newblock \titlecap{Model-based active exploration}.
\newblock \emph{\titlecap{arXiv preprint arXiv:1810.12162}}, 2018.

\bibitem[Stratonovich(1960)]{stratonovich1960hmm}
R~Stratonovich.
\newblock \titlecap{Markov's conditional processes}.
\newblock \emph{\titlecap{Teoriya Veroyatn. Primen}}, 5, 1960.

\bibitem[Sun et~al.(2017)Sun, Chen, and Carin]{sun2017bnnstructure}
S~Sun, C~Chen, L~Carin.
\newblock \titlecap{Learning structured weight uncertainty in bayesian neural
  networks}.
\newblock \emph{\titlecap{Artificial Intelligence and Statistics}}, 2017.

\bibitem[Sun et~al.(2019)Sun, Zhang, Shi, and Grosse]{sun2019fbnn}
S~Sun, G~Zhang, J~Shi, R~Grosse.
\newblock \titlecap{Functional variational bayesian neural networks}.
\newblock \emph{\titlecap{arXiv preprint arXiv:1903.05779}}, 2019.

\bibitem[Sun et~al.(2011)Sun, Gomez, and Schmidhuber]{sun2011plansurprise}
Y~Sun, F~Gomez, J~Schmidhuber.
\newblock \titlecap{Planning to be surprised: Optimal bayesian exploration in
  dynamic environments}.
\newblock \emph{\titlecap{International Conference on Artificial General
  Intelligence}}. Springer, 2011.

\bibitem[Sutton(1991)]{sutton1991dyna}
RS~Sutton.
\newblock \titlecap{Dyna, an integrated architecture for learning, planning,
  and reacting}.
\newblock \emph{\titlecap{ACM SIGART Bulletin}}, 2\penalty0 (4), 1991.

\bibitem[Sutton and Barto(2018)]{sutton2018rlbook}
RS~Sutton AG~Barto.
\newblock \emph{\titlecap{Reinforcement learning: An introduction}}.
\newblock MIT press, 2018.

\bibitem[Sutton et~al.(1999)Sutton, Precup, and Singh]{sutton1999options}
RS~Sutton, D~Precup, S~Singh.
\newblock \titlecap{Between mdps and semi-mdps: A framework for temporal
  abstraction in reinforcement learning}.
\newblock \emph{\titlecap{Artificial intelligence}}, 112\penalty0 (1-2), 1999.

\bibitem[Sutton et~al.(2000)Sutton, McAllester, Singh, and
  Mansour]{sutton2000policygradient}
RS~Sutton, DA~McAllester, SP~Singh, Y~Mansour.
\newblock \titlecap{Policy gradient methods for reinforcement learning with
  function approximation}.
\newblock \emph{\titlecap{Advances in neural information processing systems}},
  2000.

\bibitem[Teh et~al.(2017)Teh, Bapst, Czarnecki, Quan, Kirkpatrick, Hadsell,
  Heess, and Pascanu]{teh2017distral}
Y~Teh, V~Bapst, WM~Czarnecki, J~Quan, J~Kirkpatrick, R~Hadsell, N~Heess,
  R~Pascanu.
\newblock \titlecap{Distral: Robust multitask reinforcement learning}.
\newblock \emph{\titlecap{Advances in Neural Information Processing Systems}},
  2017.

\bibitem[Tirumala et~al.(2019)Tirumala, Noh, Galashov, Hasenclever, Ahuja,
  Wayne, Pascanu, Teh, and Heess]{tirumala2019maxenttasks}
D~Tirumala, H~Noh, A~Galashov, L~Hasenclever, A~Ahuja, G~Wayne, R~Pascanu,
  YW~Teh, N~Heess.
\newblock \titlecap{Exploiting hierarchy for learning and transfer in
  kl-regularized rl}.
\newblock \emph{\titlecap{arXiv preprint arXiv:1903.07438}}, 2019.

\bibitem[Todorov(2008)]{todorov2008duality}
E~Todorov.
\newblock \titlecap{General duality between optimal control and estimation}.
\newblock \emph{\titlecap{2008 47th IEEE Conference on Decision and Control}}.
  IEEE, 2008.

\bibitem[Tran et~al.(2019)Tran, Dusenberry, van~der Wilk, and
  Hafner]{tran2019bayesianlayers}
D~Tran, M~Dusenberry, M~van~der Wilk, D~Hafner.
\newblock \titlecap{Bayesian layers: A module for neural network uncertainty}.
\newblock \emph{\titlecap{Advances in Neural Information Processing Systems}},
  2019.

\bibitem[Tschannen et~al.(2019)Tschannen, Djolonga, Rubenstein, Gelly, and
  Lucic]{tschannen2019miprior}
M~Tschannen, J~Djolonga, PK~Rubenstein, S~Gelly, M~Lucic.
\newblock \titlecap{On mutual information maximization for representation
  learning}.
\newblock \emph{\titlecap{arXiv preprint arXiv:1907.13625}}, 2019.

\bibitem[Wald(1947)]{wald1947complete}
A~Wald.
\newblock \titlecap{An essentially complete class of admissible decision
  functions}.
\newblock \emph{\titlecap{The Annals of Mathematical Statistics}}, 1947.

\bibitem[Wen et~al.(2018)Wen, Vicol, Ba, Tran, and Grosse]{wen2018flipout}
Y~Wen, P~Vicol, J~Ba, D~Tran, R~Grosse.
\newblock \titlecap{Flipout: Efficient pseudo-independent weight perturbations
  on mini-batches}.
\newblock \emph{\titlecap{arXiv preprint arXiv:1803.04386}}, 2018.

\bibitem[Wiener(1948)]{wiener1948cybernetics}
N~Wiener.
\newblock \emph{\titlecap{Cybernetics or Control and Communication in the
  Animal and the Machine}}.
\newblock MIT press, 1948.

\bibitem[Williams(1992)]{williams1992reinforce}
RJ~Williams.
\newblock \titlecap{Simple statistical gradient-following algorithms for
  connectionist reinforcement learning}.
\newblock \emph{\titlecap{Machine learning}}, 8\penalty0 (3-4), 1992.

\bibitem[Williams and Peng(1991)]{williams1991maxentreinforce}
RJ~Williams J~Peng.
\newblock \titlecap{Function optimization using connectionist reinforcement
  learning algorithms}.
\newblock \emph{\titlecap{Connection Science}}, 3\penalty0 (3), 1991.

\bibitem[Xin et~al.(2020)Xin, Yu, Qin, Tang, and Zhu]{xin2020explentropy}
B~Xin, H~Yu, Y~Qin, Q~Tang, Z~Zhu.
\newblock \titlecap{Exploration entropy for reinforcement learning}.
\newblock \emph{\titlecap{Mathematical Problems in Engineering}}, 2020, 2020.

\bibitem[Yarats et~al.(2019)Yarats, Zhang, Kostrikov, Amos, Pineau, and
  Fergus]{yarats2019sacae}
D~Yarats, A~Zhang, I~Kostrikov, B~Amos, J~Pineau, R~Fergus.
\newblock \titlecap{Improving sample efficiency in model-free reinforcement
  learning from images}.
\newblock \emph{\titlecap{arXiv preprint arXiv:1910.01741}}, 2019.

\bibitem[Zhang et~al.(2018)Zhang, Sun, Duvenaud, and
  Grosse]{zhang2018naturalgradasvi}
G~Zhang, S~Sun, D~Duvenaud, R~Grosse.
\newblock \titlecap{Noisy natural gradient as variational inference}.
\newblock \emph{\titlecap{International Conference on Machine Learning}}, 2018.

\bibitem[Zhang et~al.(2019)Zhang, Vikram, Smith, Abbeel, Johnson, and
  Levine]{zhang2018solar}
M~Zhang, S~Vikram, L~Smith, P~Abbeel, M~Johnson, S~Levine.
\newblock \titlecap{Solar: deep structured representations for model-based
  reinforcement learning}.
\newblock \emph{\titlecap{International Conference on Machine Learning}}, 2019.

\bibitem[Zhao et~al.(2020)Zhao, Abbeel, and Tiomkin]{zhao2020empow}
R~Zhao, P~Abbeel, S~Tiomkin.
\newblock \titlecap{Efficient online estimation of empowerment for
  reinforcement learning}.
\newblock \emph{\titlecap{arXiv preprint arXiv:2007.07356}}, 2020.

\end{thebibliography}
\end{hyphenrules}
\clearpage
\appendix
\section{Active Inference}
\label{sec:actinf}

Divergence minimization is motivated by the free energy principle \citep{friston2010fep,friston2019physics} and its implementation active inference \citep{friston2017activeinference}. Both approaches share the interpretation of models as preferences \citep{wald1947complete,brown1981completeclass,friston2012value} and account for a variety of intrinsic objectives \citep{friston2020sophisticated}. However, typical implementations of active inference have been limited to simple tasks as of today, a problem that divergence minimization overcomes. Active inference differs from divergence minimization in the three aspects discussed below.

\paragraph{Maximizing the input probability}

Divergence minimization aims to \emph{match} the distribution of the system to the target distribution. Therefore, the agent aims to receive inputs that follow the marginal distribution of inputs under the model. In contrast, active inference aims to \emph{maximize} the probability of inputs under the model. This is often described as minimizing Bayesian surprise. Therefore, the agent aims to receive inputs that are the most probable under its model. Mathematically, this difference stems from the conditional input entropy of the actual system that distinguishes the joint KL divergence in \cref{eq:jointkl} from the expected free energy used in active inference,

\eq{
\describe{\KL[\pp(x,z) || \q(x,z)]}{joint divergence}
= \describe{\E[-\lnq(x|z)] + \EKL[\pp(z|x) || \q(z)]}{expected free energy}
- \describe{\E[-\lnpp(x)]}{input entropy}.}

Both formulations include the entropy of latent variables and thus the information gain that encourages the agent to explore informative future inputs. Moreover, in complex environments, it is unlikely that the agent ever learns everything so that its beliefs concentrate and it stops exploring. However, in this hypothetical scenario, active inference converges to the input that is most probable under its model. In contrast, divergence minimization aims to sample from the marginal input distribution under the model, resulting in a larger niche if the environment is heteroskedastic. That said, it is possible to construct a target distribution that includes the input entropy of the actual system and thus recover the expected free energy as an example of divergence minimization.

\paragraph{Expected free energy action prior}

Divergence minimization optimizes the same objective with respect to representations and actions. Therefore, actions optimize the expected information gain and representations optimize not just past accuracy but also change to support actions in maximizing the expected information gain. In contrast, active inference first optimizes the expected free energy to compute a prior over policies. After that, it optimizes the free energy with respect to both representations and actions. This means active inference optimizes the information gain only with respect to actions, without the representations changing to support better action choice based on future objective terms.

\paragraph{Bayesian model average over policies}

Typical implementations of active inference compute the action prior using a Bayesian model average. This involves computing the expected free energy for every possible policy or action sequence that is available to the agent. The action prior is then computed as the softmax over the computed values. Enumerating all policies is intractable for larger action spaces or longer planning horizons, thus limiting the applicability of active inference implementations. In contrast, divergence minimization absorbs the objective terms for action and perception into a single variational optimization thereby finessing the computational complexity of computing a separate action prior. This leads to a simple framework, allowing us to draw close connections to the deep RL literature and to scale to challenging tasks, as evidenced by the many established methods that are explained under the divergence minimization framework.
\section{KL Interpretation}
\label{sec:klinterp}

Minimizing the KL divergence has a variety of interpretations. In simple terms, it says ``optimize a function but don't be too confident.'' Decomposing \cref{eq:jointkl} shows that we maximize the expected log target while encouraging high entropy of all the random variables. Both terms are expectations under $\pp$ and thus depend on the parameter vector $\phi$,

\eq{\KL[\pp(x,z) || \q(x,z)] = \describe{\E[-\lnq(x,z)]}{energy} - \describe{\H[x,z]}{entropy}. \label{eq:energy_entropy}}

% \eq{\describe{\KL[\pp(x,z) || \q(x,z)]}{joint KL} = -\describe{\E_{p_\phi}[\lnq(x,z)]}{preferences} - \describe{\H_{p_\phi}[x,z]}{entropy}.}

The energy term expresses which system configurations we prefer. It is also known as the cross entropy loss, expected log loss, \citep{bishop2006book,murphy2012book}, energy function when unnormalized \citep{lecun2006energytutorial}, and agent preferences in control \citep{morgenstern1953seu}.

The entropy term prevents all random variables in the system from becoming deterministic, encouraging a global search over their possible values. It implements the maximum entropy principle to avoid overconfidence \citep{jaynes1957maximumentropy}, Occam's razor to prevent overfitting \citep{jefferys1992occamsrazor}, bounded rationality to halt optimization before reaching the point solution \citep{ortega2011boundedrational}, and risk-sensitivity to account for model misspecification \citep{pratt1964risk,howard1972riskrl}.

\paragraph{Expected utility}

The entropy distinguishes the KL from the expected utility objective that is typical in RL \citep{sutton2018rlbook}. Using a distribution as the optimization target is more general, as every system has a distribution but not every system has a utility function it is optimal for. Moreover, the dynamics of any stochastic system maximize only its log stationary distribution \citep{ao2013dynamicdecomp,friston2013life,ma2015mcmcdecomp}. This motivates using the desired distribution as the optimization target. Expected utility is recovered in the limit of a sharp target that outweighs the entropy.

\section{Notation}
\label{sec:background}

This section introduces notation, defines basic information-theoretic quantities, and briefly reviews KL control and variational inference for latent variable models.

\paragraph{Expectation}

A random variable $x$ represents an unknown variable that could take on one of multiple values $\bar{x}$, each with an associated probability mass or density $\p(x=\bar{x})$. Applying a function to a random variable yields a new random variable $y = f(x)$. The expectation of a random variable is the weighted average of the values it could take on, weighted by their probability,
\eq{\E[f(x)] \doteq \int f(x) \p(x) \d x.}
We use integrals here, as used for random variables that take on continuous values. For discrete variables, the integrals simplify to sums.

\paragraph{Information}

The information of an event $\bar{x}$ measures the number of bits it contains \citep{shannon1948infotheory}. Intuitively, rare events contain more information. The information is defined as the code length of the event under an optimal encoding for $x \sim \p(x)$,
\eq{\operatorname{I}(\bar{x}) \doteq \ln\left(\frac{1}{\p(\bar{x})}\right) = -\lnp(\bar{x}).}
The logarithm base 2 measures information in bits and the natural base in the unit nats.

\paragraph{Entropy}

The entropy of a random variable $x$ is the expected information of its events. It quantifies the randomness or uncertainty of the random variable. Similarly, the conditional entropy measures the uncertainty of $x$ that we expect to remain after observing another variable $y$,
\eq{\H[x] \doteq \E[-\lnp(x)], \qquad
\H[x|y] \doteq \E[-\lnp(x|y)].}
Note that the conditional entropy uses an expectation over both variables. A deterministic distribution reaches the minimum entropy of zero. The uniform distribution reaches the maximum entropy, the logarithm of the number of possible events.

\paragraph{KL divergence}

The Kullback-Leibler divergence \citep{kullback1951kl}, measures the directed similarity of one distribution to another distribution. The KL divergence is defined as the expectation under $\p$ of the log difference between the two distributions $\p$ and $\q$,
\eq{\KL[\p(x) || \q(x)] \doteq \E[\lnp(x)-\lnq(x)] = \E[-\lnq(x)] - \H[x]. \label{eq:kl}}
The KL divergence is non-negative and reaches zero if and only if $\p = \q$. Also known as relative entropy, it is the expected number of additional bits needed to describe $x$ when using the code for a different distribution $\q$ to encode events from $x \sim \p(x)$. This is shown by the decomposition as cross-entropy minus entropy shown above. Analogously to the conditional entropy, the conditional KL divergence is an expectation over both variables under the first distribution. 

\paragraph{Mutual information}

The mutual information, or simply information, between two random variables $x$ and $y$ measures how many bits the value of $x$ carries about the unobserved value of $y$. It is defined as the entropy of one variable minus its conditional entropy given the other variable,
\eq{\I[X;Y]
&\doteq \H[X]-\H[X|Y] = \E[\ln\p(x|y)-\ln\p(x)]
= \KL[\p(x,y) || \p(x)\p(y)].}
The mutual information is symmetric in its arguments and non-negative. It reaches zero if and only if $x$ and $y$ are independent so that $\p(x,y)=\p(x)\p(y)$. Intuitively, it is higher the better we can predict one variable from the other and the more random the variable is by itself. It can also be written as KL divergence between the joint and product of marginals.

\paragraph{Variational bound}

Computing the exact mutual information requires access to both the conditional and marginal distributions. When the conditional is unknown, replacing it with another distribution bounds the mutual information from below \citep{barber2003variationalinfo,poole2019bounds},
\eq{\I[x; z]
\geq \I[x; z] - \EKL[\p(x|z) || \qp(x|z)]
= \E[\lnqp(x|z)-\lnp(x)].}
Maximizing the bound with respect to the parameters $\phi$ tightens the bound, thus bringing $\qp(x|z)$ closer to $\p(x|z)$. Improving the bound through optimization gives it the name variational bound. The more flexible the family of $\qp(x|z)$, the more accurate the bound can become.

\paragraph{Dirac distribution}

The Dirac distribution \citep{dirac1958quantummechanics}, also known as point mass distribution, represents a random variable $x$ with certain event $\bar{x}$. We show an intuitive definition here; for a rigorous definition using measure theory see \citet{rudin1966analysis},
\eq{\delta_{\bar{x}}(x) \doteq
\begin{cases}
1 &\text{if } x=\bar{x} \\
0 &\text{else}.
\end{cases}}
The expectation under a Dirac distribution is simply the inner expression evaluated at the certain event, $\E<E_{\delta_{\bar{x}}(x)}>[f(x)]=f(\bar{x})$. The entropy of a Dirac distributed random variable is therefore $\smash{\H[x]=-\ln\delta_{\bar{x}}(\bar{x})=0}$ and its mutual information with another random variables is also zero.

\paragraph{KL control}

KL control \citep{todorov2008duality,kappen2009klcontrol} minimizes the KL divergence between the trajectory $x \sim \pp(x)$ of inputs $x \doteq \{x_1,x_2,\dots,x_T\}$ and a target distribution $\tau(x) \,\smash{\dotprop} \exp(r(x))$ defined with a reward $r(x)$,

\eq{\KL[\describe{\pp(x)}{trajectory} || \describe{\tau(x)}{target}]
=-\describe{\E[\lnq(x)]}{expected reward} - \describe{\H[x]}{entropy}.}

The KL between the two distributions is minimized with respect to the control rule or action sequence $\phi$, revealing the expected reward and an entropy regularizer. Because the expectations are terms of the trajectory $x$, they are integrals under its distribution $\pp$.

\paragraph{Variational inference}

Latent variable models explain inputs $x$ using latent variables $z$. They define a prior $\q(z)$ and an observation model $\q(x|z)$. To infer the posterior $\q(z|\bar{x})$ that represents a given input $\bar{x}$, we need to condition the model on the input. However, this requires inverting the observation model using Bayes rule and has no closed form in general. To overcome this intractability, variational inference \citep{hinton1993vi,jordan1999vi} optimizes a parameterized belief $\pp(z|\bar{x})$ to approximate the posterior by minimizing the KL,

\eq{\KL[\pp(z|\bar{x}) || \q(z|\bar{x})]
+ \describe{\lnq(\bar{x})}{constant}
= \describe{\KL[\pp(z|\bar{x}) || \q(z)]}{complexity}
- \describe{\E[\lnq(\bar{x}|z)]}{accuracy}.}

Adding the marginal $\q(x)$ that does not depend on $\phi$ completes the intractable posterior to the joint that can be factorized into the available parts $\q(z)$ and $\q(x|z)$. This reveals a complexity regularizer that keeps the belief close to the prior and an accuracy term that encourages the belief to be representative of the input. This objective is known as the variational free energy or ELBO.

\end{document}